\documentclass[10pt,twocolumn,letterpaper]{article}

\usepackage{graphicx}
\usepackage{subcaption}
\usepackage{float}
\usepackage[justification=raggedright]{caption}	%
\usepackage{lscape}                                         %

\usepackage[lined,ruled,linesnumbered]{algorithm2e}

\usepackage{booktabs}                   %
\usepackage{multirow}

\usepackage{paralist}
\usepackage{enumitem}

\usepackage{bm}                          %
\usepackage{epsfig}                      %
\usepackage{graphicx}                  %
\usepackage{times}
\usepackage{mathptmx}
\usepackage{mathtools}
\usepackage{amssymb,amsmath}   %

\usepackage{units}
\usepackage{color}

\usepackage{comment}

\usepackage{url}  %
\usepackage[pagebackref=true,breaklinks=true,letterpaper=true,colorlinks,bookmarks=false]{hyperref}

\usepackage{xspace}
\usepackage[table]{xcolor}
\usepackage{setspace}
\usepackage[percent]{overpic}
\usepackage{nccbbb}
\usepackage{ifthen}
\usepackage[toc,page,title]{appendix}
\usepackage{titling} %

\def\etal{et al.~}			  %
\def\eg{e.g.,~}               %
\def\ie{i.e.,~}               %
\def\etc{etc}                 %

\newlength\paramargin
\newlength\figmargin
\newlength\secmargin
\newlength\figcapmargin

\newcommand{\mpage}[2]
{
\begin{minipage}{#1\linewidth}\centering
#2
\end{minipage}
}

\newcommand{\heading}[1]
{
\vspace{1mm}
\noindent \textbf{#1}
}   

\newcommand{\secref}[1]{Section~\ref{sec:#1}}
\newcommand{\figref}[1]{Figure~\ref{fig:#1}} 
\newcommand{\tabref}[1]{Table~\ref{tab:#1}}
\newcommand{\eqnref}[1]{Equation~\ref{eq:#1}}

\long\def\ignorethis#1{}

\newcommand{\tb}[1]{\textbf{#1}}

\def\grayimg{X}                 %
\def\predimg{Y}                 %
\def\predGT{\predimg^{GT}}      %
\def\instngray{\grayimg_i}      %
\def\instnpred{\predimg_i}      %
\def\instnGT{\predimg^{GT}_i}   %
\def\instnbox{B}                %
\def\instnboxi{\instnbox_i}     %
\def\fullfetr{f^{\grayimg}_j}       %
\def\instnfetr{f^{\instngray}_j}    %
\def\resizedInstnfetr{\Bar{f^{\instngray}_j}}    %
\def\fusedfetr{f^{\tilde{\grayimg}}_j}  %
\def\fullweight{W_F}                %
\def\instnweight{W^{i}_I}           %
\def\resizedInstnweight{\Bar{W^{i}_I}}           %
\def\colorspace{$\mathrm{CIE}~\mathrm{L^{\ast}a^{\ast}b^{\ast}}$}

\graphicspath{{figure}, {images}, {example}}

\usepackage{cvpr}

\cvprfinalcopy %

\def\cvprPaperID{5819} %

\ifcvprfinal\fi

\begin{document}

\title{Instance-aware Image Colorization}

\author{
Jheng-Wei Su$^{1}$\quad\quad
Hung-Kuo Chu$^{1}$\quad\quad
Jia-Bin Huang$^{2}$\\
$^1$National Tsing Hua University
\quad\quad
$^2$Virginia Tech
\\
\url{https://ericsujw.github.io/InstColorization}
}

\twocolumn[{
\renewcommand\twocolumn[1][]{#1}
\maketitle
\begin{center}
    \centering

        \includegraphics[height=2.8cm,width=\linewidth,keepaspectratio]{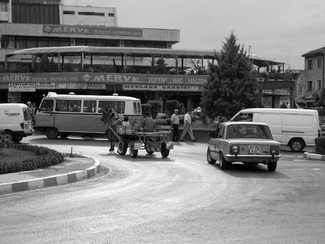} \hfill
        \includegraphics[height=2.8cm,width=\linewidth,keepaspectratio]{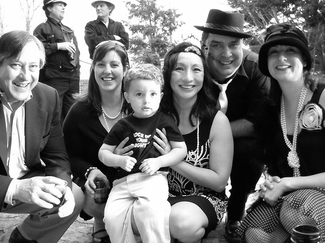} \hfill
        \includegraphics[height=2.8cm,width=\linewidth,keepaspectratio]{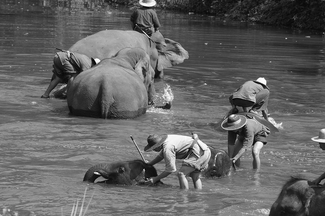} \hfill
        \includegraphics[height=2.8cm,width=\linewidth,keepaspectratio]{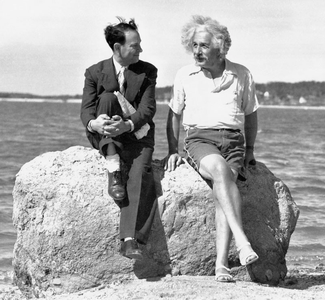} \hfill
        \includegraphics[height=2.8cm,width=\linewidth,keepaspectratio]{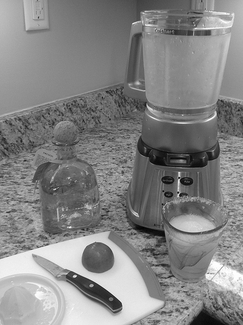}

        \includegraphics[height=2.8cm,width=\linewidth,keepaspectratio]{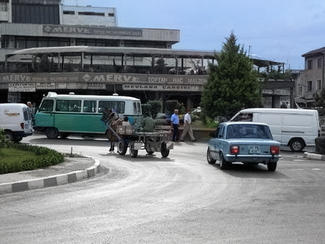} \hfill
        \includegraphics[height=2.8cm,width=\linewidth,keepaspectratio]{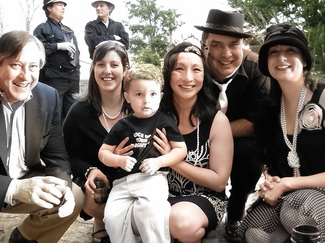} \hfill
        \includegraphics[height=2.8cm,width=\linewidth,keepaspectratio]{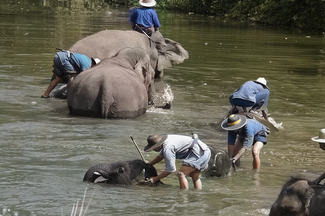} \hfill
        \includegraphics[height=2.8cm,width=\linewidth,keepaspectratio]{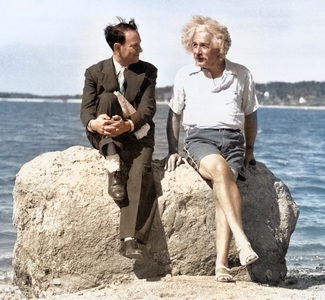} \hfill
        \includegraphics[height=2.8cm,width=\linewidth,keepaspectratio]{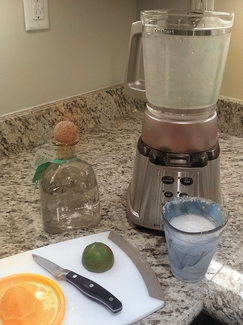}

    \vspace{\figcapmargin}
    \captionof{figure}{\tb{Instance-aware colorization.} We present an \emph{instance-aware} colorization method that is capable of producing natural and colorful results on a wide range of scenes containing multiple objects with diverse context (\eg, vehicles, people, and man-made objects).
    }
    \label{fig:teaser}
\end{center}
}]

\begin{abstract}
Image colorization is inherently an ill-posed problem with multi-modal uncertainty.
Previous methods leverage the deep neural network to map input grayscale images to plausible color outputs directly.
Although these learning-based methods have shown impressive performance, they usually fail on the input images that contain multiple objects.
The leading cause is that existing models perform learning and colorization on the entire image.
In the absence of a clear figure-ground separation, these models cannot effectively locate and learn meaningful object-level semantics.
In this paper, we propose a method for achieving instance-aware colorization.
Our network architecture leverages an off-the-shelf object detector to obtain cropped object images and uses an instance colorization network to extract object-level features.
We use a similar network to extract the full-image features and apply a fusion module to full object-level and image-level features to predict the final colors.
Both colorization networks and fusion modules are learned from a large-scale dataset.
Experimental results show that our work outperforms existing methods on different quality metrics and achieves state-of-the-art performance on image colorization. %
\end{abstract}

\section{Introduction}
\label{sec:intro}

\begin{figure*}[!t]
    \begin{subfigure}[t]{.24\linewidth}
       \includegraphics[width=\linewidth,keepaspectratio]{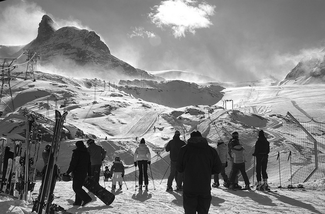}
       \includegraphics[width=\linewidth,keepaspectratio]{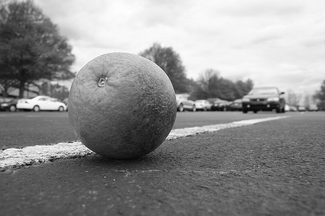}
       \caption{Input}
    \end{subfigure}
    \hfill
    \begin{subfigure}[t]{.24\linewidth}
       \includegraphics[width=\linewidth,keepaspectratio]{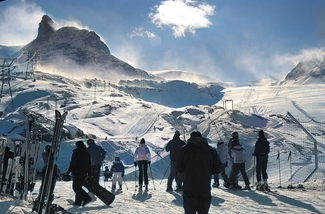}
       \includegraphics[width=\linewidth,keepaspectratio]{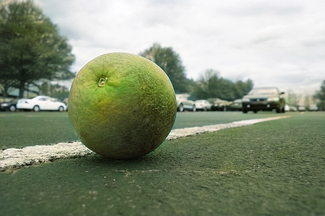}
       \caption{Deoldify~\cite{Deoldify}}
    \end{subfigure}
    \hfill
    \begin{subfigure}[t]{.24\linewidth}
       \includegraphics[width=\linewidth,keepaspectratio]{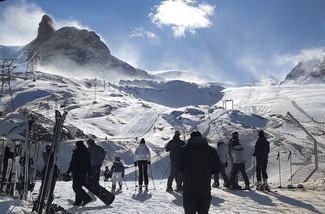}
       \includegraphics[width=\linewidth,keepaspectratio]{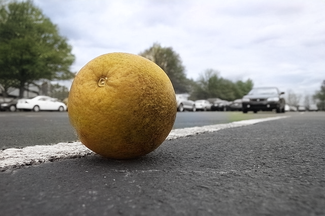}
       \caption{Zhang~\etal~\cite{Zhang-SIGGRAPH-2017}}
    \end{subfigure}
    \hfill
    \begin{subfigure}[t]{.24\linewidth}
       \includegraphics[width=\linewidth,keepaspectratio]{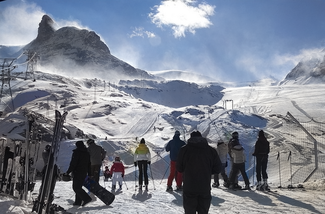}
       \includegraphics[width=\linewidth,keepaspectratio]{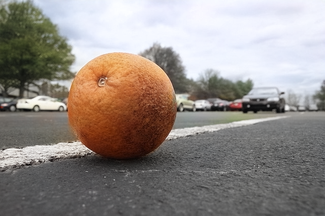}
       \caption{Ours}
    \end{subfigure}
    \vspace{\figcapmargin}
    \caption{\tb{Limitations of existing methods.}  Existing learning-based methods fail to predict plausible colors for multiple object instances such as skiers (top) and vehicles (bottom). The result of Deoldify~\cite{Deoldify}(bottom) also suffers the context confusion (biasing to green color) due to the lack of clear figure-ground separation.
    }
    \label{fig:otherbad}
\end{figure*}

Automatically converting a grayscale image to a plausible color image is an exciting research topic in computer vision and graphics, which has several practical applications such as legacy photos/video restoration or image compression.
However, predicting two missing channels from a given single-channel grayscale image is inherently an ill-posed problem.
Moreover, the colorization task could be multi-modal~\cite{Charpiat-ECCV-2008} as there are multiple plausible choices to colorize an object (\eg, a vehicle can be white, black, red, \etc).
Therefore, image colorization remains a challenging yet intriguing research problem awaiting exploration.

Traditional colorization methods rely on user intervention to provide some guidance such as color scribbles~\cite{Levin-SIGGRAPH-2004,Huang-ACMMM-2005,Yatziv-TIP-2006,Qu-SIGGRAPH-2006,Luan-EGSR-2007,Sykora-CGF-2009} or reference images~\cite{Welsh-SIGGRAPH-2002,Irony-EGSR-2005,Charpiat-ECCV-2008,Gaputa-MM-2012,Liu-SIGGRAPHASIA-2008,Chia-SIGGRAPHASIA-2011} to obtain satisfactory results. 
With the advances of deep learning, an increasing amount of efforts has focused on leveraging deep neural network and large-scale dataset such as ImageNet~\cite{ILSVRC15} or COCO-Stuff~\cite{caesar-CVPR-2018} to learn colorization in an end-to-end fashion~\cite{Cheng-ICCV-2015,Iizuka-SIGGRAPH-2016,Larsson-ECCV-2016,Zhang-ECCV-2016,Zhang-SIGGRAPH-2017,Isola-CVPR-2017,Zhao-BMVC-2018,He-SIGGRAPH-2018,Guadarrama-BMVC-2017,Royer-BMVC-2017,Deshpande-CVPR-2017,Messaoud-ECCV-2018,Deoldify}.
A variety of network architectures have been proposed to address image-level semantics~\cite{Iizuka-SIGGRAPH-2016,Larsson-ECCV-2016,Zhang-ECCV-2016,Zhao-BMVC-2018} at training or predict per-pixel color distributions to model multi-modality~\cite{Larsson-ECCV-2016,Zhang-ECCV-2016,Zhao-BMVC-2018}.
Although these learning-based methods have shown remarkable results on a wide variety of images, we observe that existing colorization models do not perform well on the images with multiple objects in a cluttered background (see~\figref{otherbad}).

In this paper, we address the above issues and propose a novel deep learning framework to achieve \emph{instance-aware} colorization.
Our key insight is that a clear figure-ground separation can dramatically improve colorization performance.
Performing colorization at the instance level is effective due to the following two reasons. 
First, unlike existing methods that learn to colorize the entire image, learning to colorize instances is a substantially easier task because it does not need to handle complex background clutter.
Second, using localized objects (\eg, from an object detector) as inputs allows the instance colorization network to learn object-level representations for accurate colorization and avoiding color confusion with the background.
Specifically, our network architecture consists of three parts:
(i) an off-the-shelf pre-trained model to detect object instances and produce cropped object images;
(ii) two backbone networks trained end-to-end for instance and full-image colorization, respectively; and
(iii) a fusion module to selectively blend features extracted from layers of the two colorization networks.
We adopt a three-step training that first trains instance network and full-image network separately, followed by training the fusion module with two backbones locked.

We validate our model on three public datasets (ImageNet~\cite{ILSVRC15}, COCO-Stuff~\cite{caesar-CVPR-2018}, and Places205~\cite{zhou-CVPR-2014}) using the network derived from Zhang~\etal~\cite{Zhang-SIGGRAPH-2017} as the backbones.
Experimental results show that our work outperforms existing colorization methods in terms of quality metrics across all datasets.
\figref{teaser} shows sample colorization results generated by our method.

Our contributions are as follows:
\begin{compactitem}
    \item A new learning-based method for fully automatic instance-aware image colorization.
    \item A novel network architecture that leverages off-the-shelf models to detect the object and learn from large-scale data to extract image features at the instance and full-image level, and to optimize the feature fusion to obtain the smooth colorization results.
    \item A comprehensive evaluation of our method on comparing with baselines and achieving state-of-the-art performance.
\end{compactitem}

\section{Related Work}
\label{sec:related}

\paragraph{Scribble-based colorization.}
Due to the multi-modal nature of image colorization problem, early attempts rely on additional high-level user scribbles (\eg, color points or strokes) to guide the colorization process~\cite{Levin-SIGGRAPH-2004,Huang-ACMMM-2005,Yatziv-TIP-2006,Qu-SIGGRAPH-2006,Luan-EGSR-2007,Sykora-CGF-2009}.
These methods, in general, formulate the colorization as a constrained optimization problem that propagates user-specified color scribbles based on some low-level similarity metrics.
For instance, Levin~\etal~\cite{Levin-SIGGRAPH-2004} encourage assigning a similar color to adjacent pixels with similar luminance.
Several follow-up approaches reduce color bleeding via edge detection~\cite{Huang-ACMMM-2005} or improve the efficiency of color propagation with texture similarity~\cite{Qu-SIGGRAPH-2006,Luan-EGSR-2007} or intrinsic distance~\cite{Yatziv-TIP-2006} .
These methods can generate convincing results with detailed and careful guidance hints provided by the user.
The process, however, is labor-intensive.
Zhang~\etal~\cite{Zhang-SIGGRAPH-2017} partially alleviate the manual efforts by combining the color hints with a deep neural network.

\paragraph{Example-based colorization.}
To reduce intensive user efforts, several works colorize the input grayscale image with the color statistics transferred from a reference image specified by the user or searched from the Internet~\cite{Welsh-SIGGRAPH-2002,Irony-EGSR-2005,Charpiat-ECCV-2008,Gaputa-MM-2012,Liu-SIGGRAPHASIA-2008,Chia-SIGGRAPHASIA-2011,He-SIGGRAPH-2018}.
These methods compute the correspondences between the reference and input image based on some low-level similarity metrics measured at pixel level~\cite{Welsh-SIGGRAPH-2002,Liu-SIGGRAPHASIA-2008}, semantic segments level~\cite{Irony-EGSR-2005,Charpiat-ECCV-2008}, or super-pixel level~\cite{Gaputa-MM-2012,Chia-SIGGRAPHASIA-2011}.
The performance of these methods is highly dependent on how similar the reference image is to the input grayscale image. 
However, finding a suitable reference image is a non-trivial task even with the aid of automatic retrieval system~\cite{Chia-SIGGRAPHASIA-2011}.
Consequently, such methods still rely on manual annotations of image regions~\cite{Irony-EGSR-2005,Chia-SIGGRAPHASIA-2011}.
To address these issues, recent advances include learning the mapping and colorization from large-scale dataset~\cite{He-SIGGRAPH-2018} and the extension to video colorization~\cite{Zhang-CVPR-2019}.

\paragraph{Learning-based colorization}
Exploiting machine learning to automate the colorization process has received increasing attention in recent years~\cite{Deshpande-ICCV-2015,Cheng-ICCV-2015,Iizuka-SIGGRAPH-2016,Larsson-ECCV-2016,Zhang-ECCV-2016,Zhang-SIGGRAPH-2017,Isola-CVPR-2017,Zhao-BMVC-2018}.
Among existing works, the deep convolutional neural network has become the mainstream approach to learn color prediction from a large-scale dataset (\eg, ImageNet~\cite{ILSVRC15}).
Various network architectures have been proposed to address two key elements for convincing colorization: semantics and multi-modality~\cite{Charpiat-ECCV-2008}.

To model semantics, Iizuka~\etal~\cite{Iizuka-SIGGRAPH-2016} and Zhao~\etal~\cite{Zhao-BMVC-2018} present a two-branch architecture that jointly learns and fuses local image features and global priors (\eg, semantic labels).
Zhang~\etal~\cite{Zhang-ECCV-2016} employ a cross-channel encoding scheme to provide semantic interpretability, which is also achieved by Larsson~\etal~\cite{Larsson-ECCV-2016} that pre-trained their network for a classification task.
To handle multi-modality, some works proposed to predict per-pixel color distributions~\cite{Larsson-ECCV-2016,Zhang-ECCV-2016,Zhao-BMVC-2018} instead of a single color.
These works have achieved impressive performance on images with moderate complexity but still suffer visual artifacts when processing complex images with multiple foreground objects as shown in~\figref{otherbad}.

Our observation is that learning semantics at either image-level~\cite{Iizuka-SIGGRAPH-2016,Zhang-ECCV-2016,Larsson-ECCV-2016} or pixel-level~\cite{Zhao-BMVC-2018} cannot sufficiently model the appearance variations of objects.
Our work thus learns object-level semantics by training on the cropped object images and then fusing the learned object-level and full-image features to improve the performance of any off-the-shelf colorization networks.

\paragraph{Colorization for visual representation learning.}
Colorization has been used as a proxy task for learning visual representation~\cite{Larsson-ECCV-2016,Zhang-ECCV-2016,larsson2017colorization,zhang2017split} and visual tracking~\cite{vondrick2018tracking}.
The learned representation through colorization has been shown to transfer well to other downstream visual recognition tasks such as image classification, object detection, and segmentation.
Our work is inspired by this line of research on self-supervised representation learning.
Instead of aiming to learn a representation that generalizes well to object detection/segmentation, we focus on leveraging the off-the-shelf pre-trained object detector to improve image colorization.

\paragraph{Instance-aware image synthesis and manipulation.} 
Instance-aware processing provides a clear figure-ground separation and facilitates synthesizing and manipulating visual appearance.
Such approaches have been successfully applied to image generation~\cite{singh2019finegan}, image-to-image translation~\cite{ma2018gan,shen2019towards,mo2018instagan}, and semantic image synthesis~\cite{wang2018high}.
Our work leverages a similar high-level idea with these methods but differs in the following three aspects.
First, unlike DA-GAN~\cite{ma2018gan} and FineGAN~\cite{singh2019finegan} that focus only on one \emph{single} instance, our method is capable of handling complex scenes with \emph{multiple} instances via the proposed feature fusion module.
Second, in contrast to \emph{Insta}GAN~\cite{mo2018instagan} that processes \emph{non-overlapping} instances \emph{sequentially}, our method considers all \emph{potentially overlapping} instances \emph{simultaneously} and produces spatially coherent colorization.
Third, compared with Pix2PixHD~\cite{wang2018high} that uses instance boundary for improving synthesis quality, our work uses \emph{learned} weight maps for blending features from multiple instances.

\begin{figure*}[ht]
\begin{overpic}[width=\textwidth]{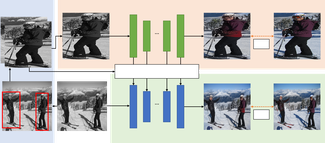}
\put(1, 41){\begin{minipage}{0.15\textwidth}\centering Object Detection\\(\secref{objectdetection}) \end{minipage}}
\put(1.5, 21){\begin{minipage}{0.1\textwidth}\centering $\{\instngray\}_{i=1}^N$ \end{minipage}}
\put(2, 1.5){\begin{minipage}{0.1\textwidth}\centering $\instnboxi$ \end{minipage}}
\put(21.5, 23.5){\begin{minipage}{0.1\textwidth}\centering $\instngray$ \end{minipage}}
\put(19.5, 1.5){\begin{minipage}{0.1\textwidth}\centering Input $\grayimg$ \end{minipage}}
\put(38, 41){\begin{minipage}{0.2\textwidth}\centering Instance Colorization\\(\secref{backbone}) \end{minipage}}
\put(36, 21.5){\begin{minipage}{0.25\textwidth}\centering Fusion Module (\secref{fusionmodule}) \end{minipage}}
\put(38, 2){\begin{minipage}{0.2\textwidth}\centering (\secref{backbone})\\Full-image Colorization \end{minipage}}
\put(65, 23.5){\begin{minipage}{0.1\textwidth}\centering $\instnpred$ \end{minipage}}
\put(87, 23.5){\begin{minipage}{0.1\textwidth}\centering $\instnGT$ \end{minipage}}
\put(65, 1.5){\begin{minipage}{0.1\textwidth}\centering $\predimg$ \end{minipage}}
\put(87, 1.5){\begin{minipage}{0.1\textwidth}\centering $\predGT$ \end{minipage}}
\put(75.5, 30.5){\begin{minipage}{0.1\textwidth}\centering loss \end{minipage}}
\put(75.5, 8.5){\begin{minipage}{0.1\textwidth}\centering loss \end{minipage}}
\end{overpic}
\caption{\tb{Method overview.} Given a grayscale image $\grayimg$ as input, our model starts with detecting the object bounding boxes ($\instnboxi$) using an off-the-shelf object detection model. We then crop out every detected instance $\instngray$ via $\instnboxi$ and use instance colorization network to colorize $\instngray$. However, as the instances' colors may not be compatible with respect to the predicted background colors, we propose to fuse all the instances' feature maps in every layer with the extracted full-image feature map using the proposed fusion module. We can thus obtain globally consistent colorization results $\predimg$. Our training process sequentially trains our full-image colorization network, and the instance colorization network, and the proposed fusion module.}
\label{fig:overview}
\end{figure*}
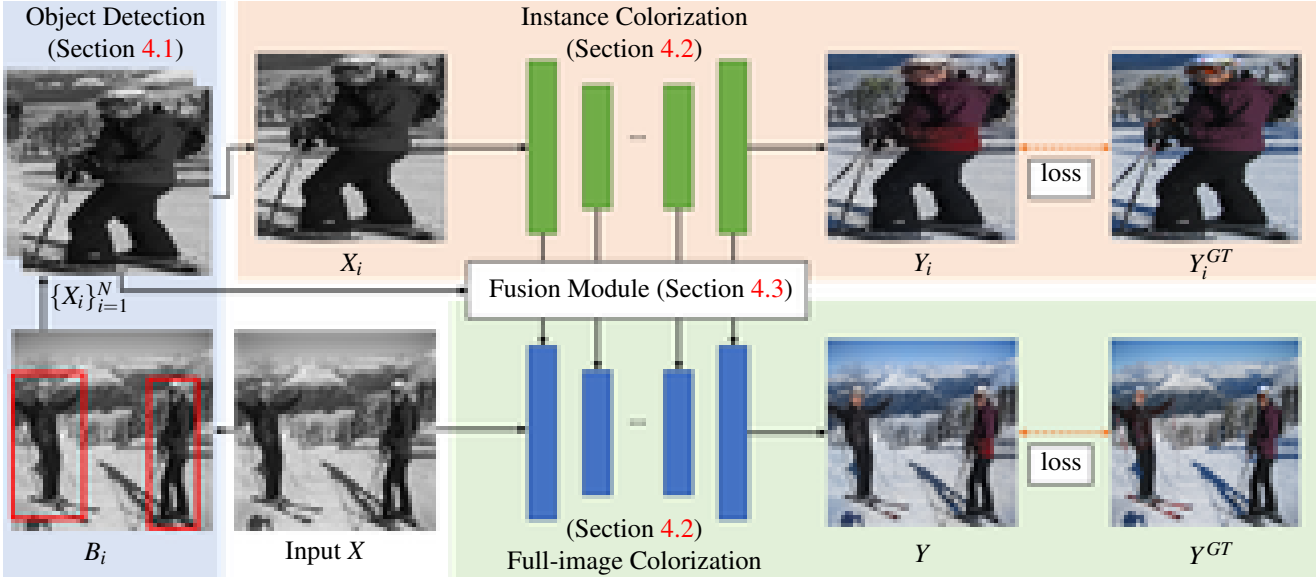

\section{Overview}
\label{sec:overview}

Our system takes a grayscale image $\grayimg\in\mathbb{R}^{H\times W\times 1}$ as input and predicts its two missing color channels $\predimg\in\mathbb{R}^{H\times W\times 2}$ in the {\colorspace} color space in an end-to-end fashion.
\figref{overview} illustrates our network architecture.
First, we leverage an off-the-shelf pre-trained object detector to obtain multiple object bounding boxes $\{\instnboxi\}^N_{i=1}$ from the grayscale image, where N is the number of instances.
We then generate a set of instance images $\{\instngray\}^N_{i=1}$ by resizing the images cropped from the grayscale image using the detected bounding boxes (\secref{objectdetection}).
Next, we feed each instance image $\instngray$ and input grayscale image $\grayimg$ to the {\em instance colorization network} and {\em full-image colorization network}, respectively.
The two networks share the same architecture (but different weights). 
We denote the extracted feature map of instance image $\instngray$ and grayscale image $\grayimg$ at the $j$-th network layer as $\instnfetr$ and $\fullfetr$ (\secref{backbone}).
Finally, we employ a {\em fusion module} that fuses all the instance features $\{\instnfetr\}^N_{i=1}$ with the full-image feature $\fullfetr$ at each layer. The fused full image feature at $j$-th layer, denoted as $\fusedfetr$, is then fed forward to $j+1$-th layer. 
This step repeats until the last layer and obtains the predict color image $\predimg$ (\secref{fusionmodule}).
We adopt a sequential approach that first trains the full-image network, followed by the instance network, and finally trains the feature fusion module by freezing the above two networks.

\section{Method}
\label{sec:method}

\subsection{Object detection}
\label{sec:objectdetection}
Our method leverages detected object instances for improving image colorization.
To this end, we employ an off-the-shelf pre-trained network, Mask R-CNN~\cite{He-ICCV-2017}, as our object detector.
After detecting each object's bounding box $\instnboxi$, we crop out corresponding grayscale instance image $\instngray$ and color instance image $\instnGT$ from $\grayimg$ and $\predGT$, and resize the cropped images to a resolution of $256\times256$.

\begin{figure*}[t]
\begin{overpic}[width=\textwidth]{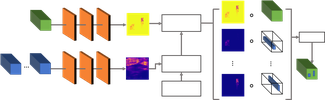}
\put(0, 29){\begin{minipage}{0.2\textwidth}\small \centering Full-image Feature ($\fullfetr$) \end{minipage}}
\put(0, 5){\begin{minipage}{0.2\textwidth}\small \centering Instance Feature ($\{\instnfetr\}^N_{i=1}$) \end{minipage}}
\put(35, 16){\begin{minipage}{0.15\textwidth}\small \centering Full-image Weight Map ($\fullweight$) \end{minipage}}
\put(35, 3){\begin{minipage}{0.15\textwidth}\small \centering Instance Weight Map ($\instnweight$) \end{minipage}}
\put(50.5, 3){\begin{minipage}{0.1\textwidth}\small \centering Bounding Boxes ($\instnboxi$) \end{minipage}}
\put(50.5, 10.5){\begin{minipage}{0.11\textwidth}\small \centering Resize and Zero Padding \end{minipage}}
\put(50.5, 23.5){\begin{minipage}{0.11\textwidth}\small \centering Softmax Normalization \end{minipage}}
\put(90.5, 18.5){\begin{minipage}{0.11\textwidth}\small \centering Sum Up \end{minipage}}
\put(90, 2.5){\begin{minipage}{0.12\textwidth}\small \centering Fused Feature ($\fusedfetr$) \end{minipage}}
\end{overpic}
\caption{\tb{Feature fusion module.} Given the full-image feature $\fullfetr$ and a bunch of instance features $\{\instnfetr\}^N_{i=1}$ from the $j$-th layer of the colorization network, we first predict the corresponding weight map $\fullweight$ and $\instnweight$ through a small neural network with three convolutional layers. Both instance feature and weight map are resized, padded with zero to match the original size and local in the full image. The final fused feature $\fusedfetr$ is thus computed using the weighted sum of retargeted features (see~\eqnref{fusion}).
} 
\label{fig:fusion}
\end{figure*}
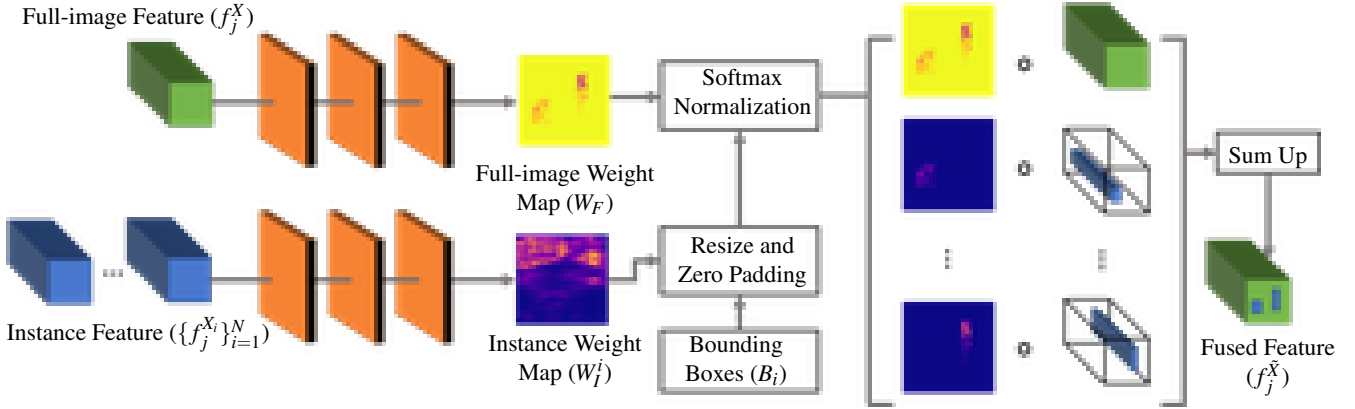

\subsection{Image colorization backbone}
\label{sec:backbone}
As shown in~\figref{overview}, our network architecture contains two branches of colorization networks, one for colorizing the instance images and the other for colorizing the full image.
We choose the architectures of these two networks so that they have the same number of layers to facilitate feature fusion (discussed in the next section). 
In this work, we adopt the main colorization network introduced in Zhang~\etal~\cite{Zhang-SIGGRAPH-2017} as our backbones.
Although these two colorization networks alone could predict the color instance images $\instnpred$ and full image $\predimg$, we found that a na\"ive blending of these results yield visible visual artifacts due to the inconsistency of the overlapping pixels.
In the following section, we elaborate on how to fuse the intermediate feature maps from both instance and full-image networks to produce accurate and coherent colorization.

\subsection{Fusion module}
\label{sec:fusionmodule}
Here, we discuss how to fuse the full-image feature with multiple instance features to achieve better colorization.
\figref{fusion} shows the architecture of our fusion module.
Since the fusion takes place at multiple layers of the colorization networks, for the sake of simplicity, we only present the {\em fusion module} at the $j$-th layer.
Apply the module to all the other layers is straightforward. 

The fusion module takes inputs: (1) a full-image feature $\fullfetr$; (2) a bunch of instance features and corresponding object bounding boxes $\{\instnfetr,\instnboxi\}^N_{i=1}$.
For both kinds of features, we devise a small neural network with three convolutional layers to predict full-image weight map $\fullweight$ and per-instance weight map $\instnweight$.
To fuse per-instance feature $\instnfetr$ to the full-image feature $\fullfetr$, we utilize the input bounding box $\instnboxi$, which defines the size and location of the instance.
Specifically, we resize the instance feature $\instnfetr$ as well as the weight map $\instnweight$ to match the size of full-image and do zero padding on both of them. 
We denote resized the instance feature and weight map as $\resizedInstnfetr$ and $\resizedInstnweight$.
After that, we stack all the weight maps, apply softmax on each pixel, and obtain the fused feature using a weighted sum as follows:
\begin{equation}
    \label{eq:fusion}
    \fusedfetr = \fullfetr \circ \fullweight + \sum^N_{i=1} \resizedInstnfetr \circ \resizedInstnweight,
\end{equation}
where $N$ is the number of instances.

\subsection{Loss Function and Training}
\label{sec:loss}
Following Zhang~\etal~\cite{Zhang-SIGGRAPH-2017}, we adopt the smooth-$\ell_1$ loss with $\delta=1$ as follows:
\begin{equation}
\label{eq:loss}
\resizebox{0.9\hsize}{!}{
$\ell_{\delta}(x, y) = \frac{1}{2}(x - y)^{2}\bbbone_{\{|x - y|<\delta\}}+  \delta(|x - y|-\frac{1}{2}\delta)\bbbone_{\{|x - y|\geqslant\delta\}}$
}    
\end{equation}
We train the whole network sequentially as follows.
First, we train the full-image colorization and transfer the learned weights to initialize the instance colorization network.
We then train the instance colorization network.
Lastly, we freeze the weights in both the full-image model and instance model and move on training the fusion module.

\begin{table*}[t!]
    \centering
    \caption{\textbf{Quantitative comparison at the full-image level}. The methods in the first block are trained using the ImageNet dataset. The symbol $*$ denotes the methods that are finetuned on the COCO-Stuff training set.}
    \resizebox{\textwidth}{!}{
        \begin{tabular}{@{}lclllclllclll@{}}
        \toprule
        \multirow{2}{*}{Method} & \phantom{abc} & 
        \multicolumn{3}{c}{Imagenet ctest10k} & \phantom{abc} & 
        \multicolumn{3}{c}{COCOStuff validation split} &\phantom{abc} & 
        \multicolumn{3}{c}{Places205 validation split}\\
        \cmidrule{3-5} \cmidrule{7-9} \cmidrule{11-13}
        && $LPIPS\downarrow$ & $PSNR\uparrow$ & $SSIM\uparrow$ && $LPIPS\downarrow$ & $PSNR\uparrow$ & $SSIM\uparrow$ && $LPIPS\downarrow$ & $PSNR\uparrow$ & $SSIM\uparrow$\\
        \midrule
        lizuka~\etal~\cite{Iizuka-SIGGRAPH-2016} && 0.200 & 23.636 & 0.917 && 0.185 & 23.863 & 0.922 && 0.146 & 25.581 & 0.950 \\
        Larsson~\etal~\cite{Larsson-ECCV-2016} && 0.188 & 25.107 & 0.927 && 0.183 & 25.061 & 0.930 && 0.161 & 25.722 & 0.951 \\
        Zhang~\etal~\cite{Zhang-ECCV-2016} && 0.238 & 21.791 & 0.892 && 0.234 & 21.838 & 0.895 && 0.205 & 22.581 & 0.921 \\
        Zhang~\etal~\cite{Zhang-SIGGRAPH-2017} && 0.145 & 26.166 & 0.932 && 0.138 & 26.823 & 0.937 && 0.149 & 25.823 & 0.948 \\
        Deoldify~\etal~\cite{Deoldify} && 0.187 & 23.537 & 0.914 && 0.180 & 23.692 & 0.920 && 0.161 & 23.983 & 0.939 \\
        Lei~\etal~\cite{Lei-CVPR-2019} && 0.202 & 24.522 & 0.917 && 0.191 & 24.588 & 0.922 && 0.175 & 25.072 & 0.942 \\
        Ours && \textbf{0.134} & \textbf{26.980} & \textbf{0.933} && \textbf{0.125} & \textbf{27.777} & \textbf{0.940} && \textbf{0.130} & \textbf{27.167} & \textbf{0.954}\\
        \midrule
        Zhang~\etal~\cite{Zhang-SIGGRAPH-2017}* && 0.140 & 26.482 & 0.932 && 0.128 & 27.251 & 0.938 && 0.153 & 25.720 & 0.947 \\
        Ours* && \textbf{0.125} & \textbf{27.562} & \textbf{0.937} && \textbf{0.110} & \textbf{28.592} & \textbf{0.944} && \textbf{0.120} & \textbf{27.800} & \textbf{0.957}\\
        \bottomrule
        \end{tabular}
    }
    \label{tab:comparision}
\end{table*}

\section{Experiments}
\label{sec:results}

In this section, we present extensive experimental results to validate the proposed instance-aware colorization algorithm. 
We start by describing the datasets used in our experiments, performance evaluation metrics, and implementation details (\secref{setting}).
We then report the quantitative evaluation of three large-scale datasets and compare our results with the state-of-the-art colorization methods (\secref{quantitative}).
We show sample colorization results on several challenging images (\secref{visual}).
We carry out three ablation studies to validate our design choices (\secref{ablation}).
Beyond standard performance benchmarking, we demonstrate the application of colorizing legacy black and white photographs (\secref{legacy}).
We conclude the section with examples where our method fails (\secref{failure}).
Please refer to the project webpage for the dataset, source code, and additional visual comparison.

\subsection{Experimental setting}
\label{sec:setting}

\heading{Datasets.}
We use three datasets for training and evaluation.

\emph{ImageNet}~\cite{ILSVRC15}: ImageNet dataset has been used by many existing colorization methods as a benchmark for performance evaluation. We use the original training split (~1.3 million images) for training all the models and use the testing split (ctest10k) provided by~\cite{Larsson-ECCV-2016} with 10,000 images for evaluation.

\emph{COCO-Stuff}~\cite{caesar-CVPR-2018}: In contrast to the \emph{object-centric} images in the ImageNet dataset, the COCO-Stuff dataset contains a wide variety of natural scenes with multiple objects present in the image. 
There are 118K images (each image is associated with a bounding box, instance segmentation, and semantic segmentation annotations). 
We use the 5,000 images in the original validation set for evaluation.

\emph{Places205}~\cite{zhou-CVPR-2014}: To investigate how well a colorization method performs on images from a different dataset, we use the 20,500 testing images (from 205 categories) from the Places205 for evaluation. 
Note that we use the Place205 dataset only for evaluating the transferability. 
We do not use its training set and the scene category labels for training.

\heading{Evaluation metrics.}
Following the experimental protocol by existing colorization methods, we report the PSNR and SSIM to quantify the colorization quality.
To compute the SSIM on color images, we average the SSIM values computed from individual channels.
We further use the recently proposed perceptual metric LPIPS by Zhang~\etal~\cite{zhang-CVPR-2018} (version 0.1; with VGG backbone).

\heading{Training details.}
We adopt a three-step training process on the ImageNet dataset as follows.

(1) \emph{Full-image colorization network}: We initialize the network with the pre-trained weight provided by \cite{Zhang-SIGGRAPH-2017}. We train the network for two epochs with a learning rate of 1e-5. 
(2) \emph{Instance colorization network}: We start with the pre-trained weight from the trained full-image colorization network above and finetune the model for five epochs with a learning rate of 5e-5 on the extracted instances from the dataset. 
(3) \emph{Fusion module}: Once both the full-image and instance network have been trained (\ie, warmed-up), we integrate them with the proposed fusion module. 
We finetune all the trainable parameters for 2 epochs with a learning rate of 2e-5. 
In our implementation, the numbers of channels of full-image feature, instance feature and fused feature in all 13 layers are 64, 128, 256, 512, 512, 512, 512, 256, 256, 128, 128, 128 and 128.

In all the training processes, we use the ADAM optimizer~\cite{kingma2014adam} with $\beta_1 = 0.99$ and $\beta_2= 0.999$. 
For training, we resize all the images to a resolution of $256\times256$. 
Training the model on the ImageNet takes about three days on a desktop machine with one single RTX 2080Ti GPU.

\begin{figure*}[!t]
    \begin{subfigure}[t]{.16\linewidth}
       \includegraphics[width=\linewidth,keepaspectratio]{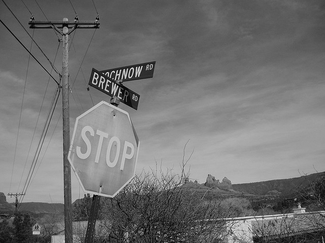}
      \includegraphics[width=\linewidth,keepaspectratio]{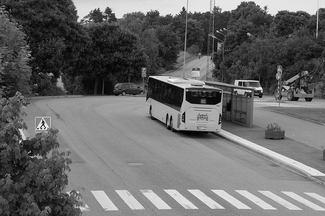}
       \includegraphics[width=\linewidth,keepaspectratio]{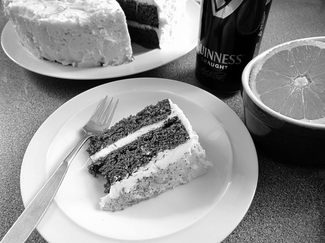}
      \includegraphics[width=\linewidth,keepaspectratio]{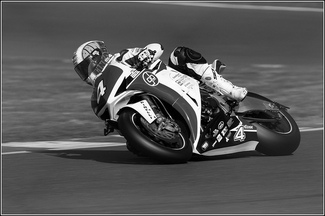}
       \includegraphics[width=\linewidth,keepaspectratio]{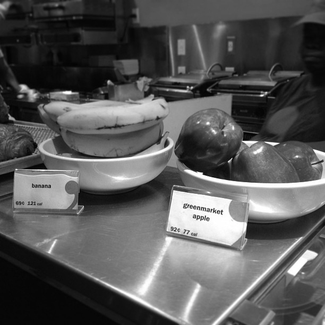}
       \caption{Input}
    \end{subfigure}
    \hfill
    \begin{subfigure}[t]{.16\linewidth}
       \includegraphics[width=\linewidth,keepaspectratio]{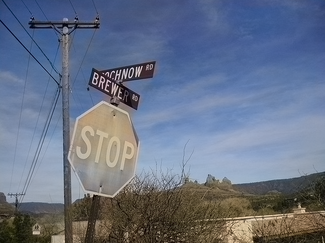}
      \includegraphics[width=\linewidth,keepaspectratio]{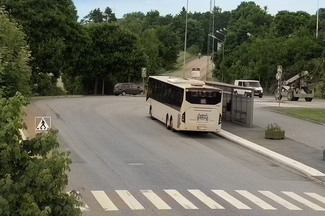}
       \includegraphics[width=\linewidth,keepaspectratio]{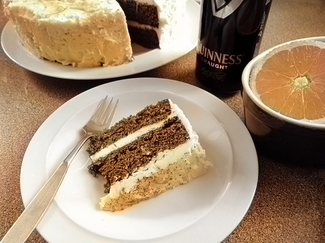}
      \includegraphics[width=\linewidth,keepaspectratio]{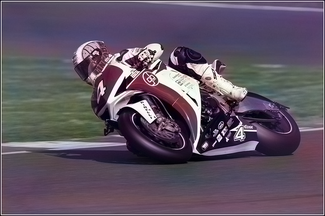}
       \includegraphics[width=\linewidth,keepaspectratio]{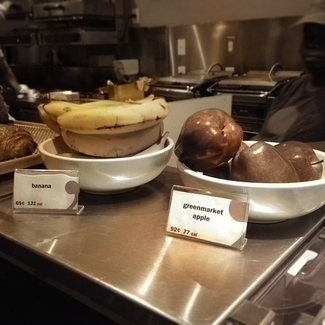}
       \caption{Iizuka~\etal~\cite{Iizuka-SIGGRAPH-2016}}
    \end{subfigure}
    \hfill
    \begin{subfigure}[t]{.16\linewidth}
       \includegraphics[width=\linewidth,keepaspectratio]{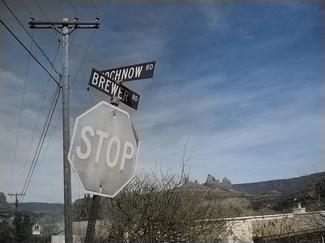}
      \includegraphics[width=\linewidth,keepaspectratio]{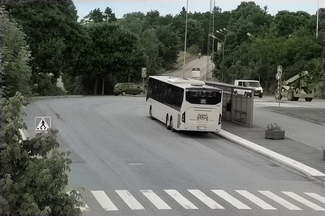}
       \includegraphics[width=\linewidth,keepaspectratio]{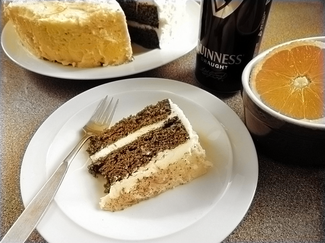}
      \includegraphics[width=\linewidth,keepaspectratio]{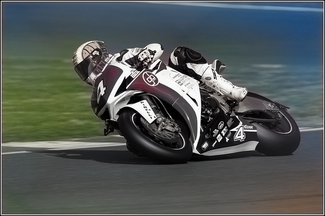}
       \includegraphics[width=\linewidth,keepaspectratio]{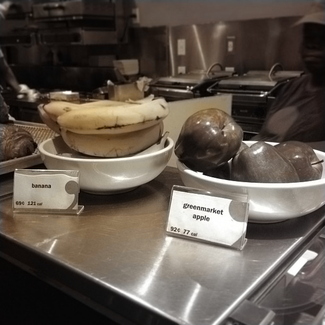}
       \caption{Larrson~\etal~\cite{larsson2017colorization}}
    \end{subfigure}
    \hfill
    \begin{subfigure}[t]{.16\linewidth}
       \includegraphics[width=\linewidth,keepaspectratio]{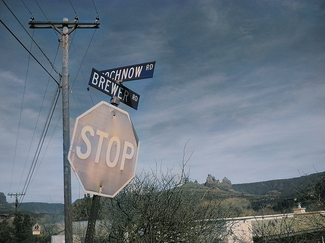}
      \includegraphics[width=\linewidth,keepaspectratio]{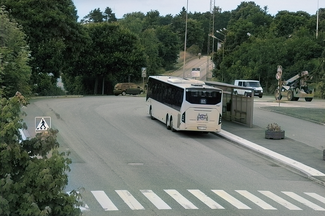}
       \includegraphics[width=\linewidth,keepaspectratio]{figure/images/visual_compare/000000496954_deoldify.png}
      \includegraphics[width=\linewidth,keepaspectratio]{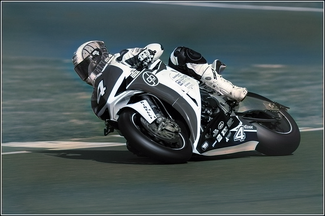}
       \includegraphics[width=\linewidth,keepaspectratio]{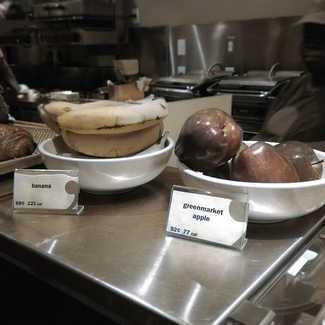}
       \caption{Deoldify~\cite{Deoldify}}
    \end{subfigure}
    \hfill
    \begin{subfigure}[t]{.16\linewidth}
       \includegraphics[width=\linewidth,keepaspectratio]{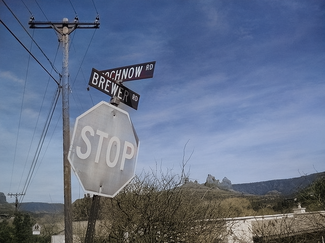}
      \includegraphics[width=\linewidth,keepaspectratio]{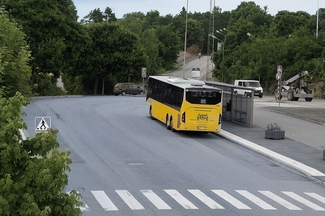}
       \includegraphics[width=\linewidth,keepaspectratio]{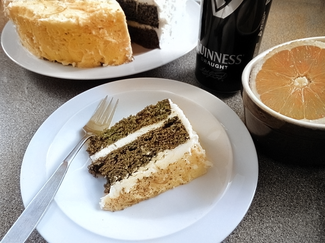}
      \includegraphics[width=\linewidth,keepaspectratio]{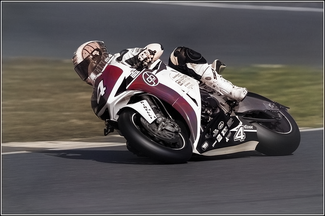}
       \includegraphics[width=\linewidth,keepaspectratio]{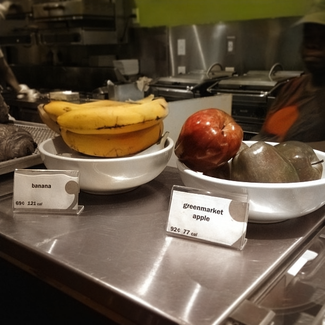}
       \caption{Zhang~\etal~\cite{Zhang-SIGGRAPH-2017}}
    \end{subfigure}
    \hfill
    \begin{subfigure}[t]{.16\linewidth}
       \includegraphics[width=\linewidth,keepaspectratio]{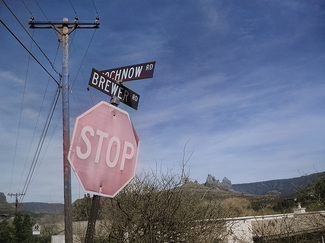}
      \includegraphics[width=\linewidth,keepaspectratio]{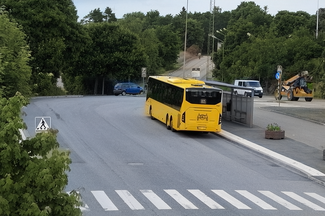}
       \includegraphics[width=\linewidth,keepaspectratio]{figure/images/visual_compare/000000496954_ours.png}
      \includegraphics[width=\linewidth,keepaspectratio]{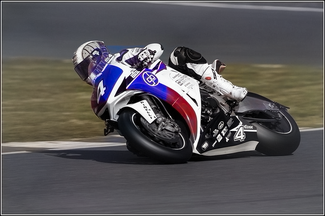}
       \includegraphics[width=\linewidth,keepaspectratio]{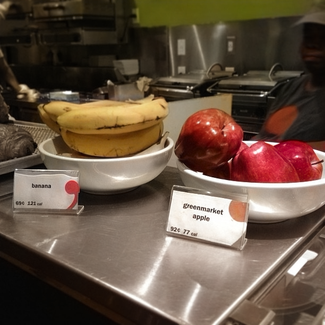}
       \caption{Ours}
    \end{subfigure}
    \caption{\tb{Visual Comparisons with the state-of-the-arts.} Our method predicts visually pleasing colors from complex scenes with multiple object instances.}
    \label{fig:comparison}
\end{figure*}

\subsection{Quantitative comparisons}\label{sec:quantitative}
\heading{Comparisons with the state-of-the-arts.}
We report the quantitative comparisons on three datasets in \tabref{comparision}. 
The first block of the results shows models trained on the ImageNet dataset.
Our instance-aware model performs favorably against several recent methods~\cite{Iizuka-SIGGRAPH-2016, larsson2017colorization, Zhang-ECCV-2016, Zhang-SIGGRAPH-2017, Deoldify, Lei-CVPR-2019} on all three datasets, highlighting the effectiveness of our approach.
Note that we adopted the automatic version of Zhang~\etal~\cite{Zhang-SIGGRAPH-2017} (\ie, without using any color guidance) in all the experiments.
In the second block, we show the results using our model finetuned on the COCO-Stuff training set (denoted by the ``*").
As the COCO-Stuff dataset contains more diverse and challenging scenes, our results show that finetuning on the COCO-Stuff dataset further improves the performance on the \emph{other two datasets} as well.
To highlight the effectiveness of the proposed instance-aware colorization module, we also report the results of Zhang~\etal~\cite{Zhang-SIGGRAPH-2017} finetuned on the same dataset as a strong baseline for a fair comparison.
For evaluating the performance at the \emph{instance-level}, we take the full-image ground-truth/prediction and crop the instances using the ground-truth bounding boxes to form instance-level ground-truth/predictions.
\tabref{instance_comp} summarizes the performance computed by averaging over all the instances on the COCO-Stuff dataset.
The results present a significant performance boost gained by our method in all metrics, which further highlights the contribution of instance-aware colorization to the improved performance.

\begin{table}[!t]
    \centering
    \caption{\textbf{Quantitative comparison at the instance level.} The methods in the first block are trained using the ImageNet dataset. The symbol $*$ denotes the methods that are finetuned on the COCO-Stuff training set.}
    \resizebox{.45\textwidth}{!}{
        \begin{tabular}{@{}lclll@{}}
            \toprule
            \multirow{2}{*}{Method} & \phantom{abc} & 
            \multicolumn{3}{c}{COCOStuff validation split} \\
            \cmidrule{3-5}
            && $LPIPS\downarrow$ & $PSNR\uparrow$ & $SSIM\uparrow$ \\
            \midrule
            lizuka~\etal~\cite{Iizuka-SIGGRAPH-2016} && 0.192 & 23.444 & 0.900 \\
            Larsson~\etal~\cite{Larsson-ECCV-2016} && 0.179 & 25.249 & 0.914 \\
            Zhang~\etal~\cite{Zhang-ECCV-2016} && 0.219 & 22.213 & 0.877 \\
            Zhang~\etal~\cite{Zhang-SIGGRAPH-2017} && 0.154 & 26.447 & 0.918 \\
            Deoldify~\etal~\cite{Deoldify} && 0.174 & 23.923 & 0.904 \\
            Lei~\etal~\cite{Lei-CVPR-2019} && 0.177 & 24.914 & 0.908 \\
            Ours && \textbf{0.115} & \textbf{28.339} & \textbf{0.929} \\
            \midrule
            Zhang~\etal~\cite{Zhang-SIGGRAPH-2017}* && 0.149 & 26.675 & 0.919 \\
            Ours* && \textbf{0.095} & \textbf{29.522} & \textbf{0.938} \\
            \bottomrule
        \end{tabular}
    }
    \label{tab:instance_comp}
\end{table}

\heading{User study.} 
We conducted a user study to quantify the user-preference on the colorization results generated by our method and another two strong baselines, Zhang~\etal~\cite{Zhang-SIGGRAPHASIA-2018} (finetuned on the COCO-Stuff dataset) and a popular online colorization method DeOldify~\cite{Deoldify}.
We randomly select 100 images from the \emph{COCO-Stuff} validation dataset.
For each participant, we show him/her a pair of colorized results and ask for the preference (forced-choice comparison).
In total, we have 24 participants casting 2400 votes in total.
The results show that on average our method is preferred when compared with Zhang~\etal~\cite{Zhang-SIGGRAPHASIA-2018} ($61\%$ v.s. $39\%$) and DeOldify~\cite{Deoldify} ($72\%$ v.s. $28\%$). 
Interestingly, while DeOldify does not produce accurate colorization evaluated in the benchmark experiment, the saturated colorized results are sometimes more preferred by the users.

\subsection{Visual results}
\label{sec:visual}

\heading{Comparisons with the state-of-the-art.}
\figref{comparison} shows sample comparisons with other competing baseline methods on \emph{COCO-Stuff}. 
In general, we observe a consistent improvement in visual quality, particularly for scenes with multiple instances.

\begin{figure}[!t]
    \begin{subfigure}[!t]{.39\linewidth}\centering
       \includegraphics[width=\linewidth,keepaspectratio]{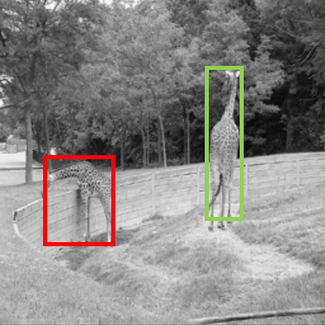}\\
       {Input}
    \end{subfigure}
    \begin{subfigure}[!t]{.19\linewidth}\centering
       \includegraphics[width=\linewidth,keepaspectratio]{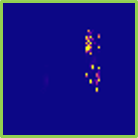}
       \includegraphics[width=\linewidth,keepaspectratio]{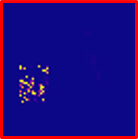}
       {Layer3}
    \end{subfigure}
    \begin{subfigure}[!t]{.19\linewidth}\centering
       \includegraphics[width=\linewidth,keepaspectratio]{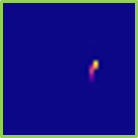}
       \includegraphics[width=\linewidth,keepaspectratio]{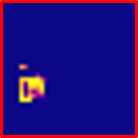}\\
       {Layer7}
    \end{subfigure}
    \begin{subfigure}[!t]{.19\linewidth}\centering
       \includegraphics[width=\linewidth,keepaspectratio]{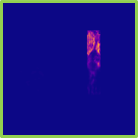}
       \includegraphics[width=\linewidth,keepaspectratio]{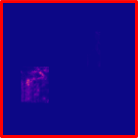}\\
       {Layer10}
    \end{subfigure}
    \\
    \begin{subfigure}[!t]{0.49\linewidth}\centering
        \includegraphics[width=\linewidth,keepaspectratio]{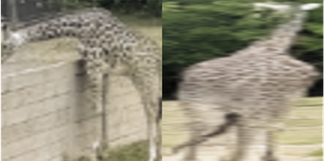}
        {Zhang~\etal~\cite{Zhang-SIGGRAPH-2017}}
    \end{subfigure}
    \begin{subfigure}[!t]{0.49\linewidth}\centering
        \includegraphics[width=\linewidth,keepaspectratio]{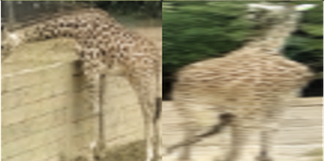}
        {Our results}
    \end{subfigure}
    \vspace{\figcapmargin}
    \caption{\tb{Visualizing the fusion network.} The visualized weighted mask in layer3, layer7 and layer10 show that our model learns to \emph{adaptively} blend the features across different layers. Fusing instance-level features help improve colorization.}
    \label{fig:vismask}
\end{figure}

\heading{Visualizing the fusion network.} \figref{vismask} visualizes the learned masks for fusing instance-level and full-image level features at multiple levels. 
We show that the proposed instance-aware processing leads to improved visual quality for complex scenarios.

\begin{table*}[t!]
\caption{\tb{Ablations.} We validate our design choices by comparing with several alternative options.}
\label{tab:ablation}
    \begin{subtable}[t]{0.325\textwidth}
        \centering
        \caption{Different Fusion Part}
        \resizebox{\textwidth}{!}{
            \begin{tabular}{@{}ccclll@{}}
                \toprule
                \multicolumn{2}{c}{Fusion Part} & \phantom{abc} & 
                \multicolumn{3}{c}{COCOStuff validation split}\\
                \cmidrule{1-2} \cmidrule{4-6}
                Encoder & Decoder && $LPIPS\downarrow$ & $PSNR\uparrow$ & $SSIM\uparrow$ \\ \midrule
                $\times$ & $\times$ && 0.128 & 27.251 & 0.938 \\
                \checkmark & $\times$ && 0.120 & 28.146 & 0.942 \\
                $\times$ & \checkmark && 0.117 & 27.959 & 0.941 \\
                \checkmark & \checkmark && \textbf{0.110} & \textbf{28.592} & \textbf{0.944} \\
                \bottomrule
            \end{tabular}
        }
        \label{tab:ablation_fusion}
    \end{subtable}
    \hfill
    \begin{subtable}[t]{0.325\textwidth}
        \centering
        \caption{Different Bounding Box Selection}
        \resizebox{\textwidth}{!}{
            \begin{tabular}{@{}lclll@{}}
                \toprule
                \multirow{2}{*}{Box Selection} & \phantom{abc} & 
                \multicolumn{3}{c}{COCOStuff validation split} \\
                \cmidrule{3-5}
                && $LPIPS\downarrow$ & $PSNR\uparrow$ & $SSIM\uparrow$ \\
                \midrule
                Select top 8 && \textbf{0.110} & \textbf{28.592} & \textbf{0.944} \\
                Random select 8 && 0.113 & 28.386 & 0.943 \\
                Select by threshold && 0.117 & 28.139 & 0.942 \\
                G.T. bounding box && 0.111 & 28.470 & \textbf{0.944} \\
                \bottomrule
            \end{tabular}
        }
        \label{tab:ablation_box}
    \end{subtable}
    \hfill
    \begin{subtable}[t]{0.325\textwidth}
        \centering
        \caption{Different Weighted Sum}
        \resizebox{\textwidth}{!}{
            \begin{tabular}{@{}lclll@{}}
                \toprule
                \multirow{2}{*}{Weighted Sum} & \phantom{abc} & 
                \multicolumn{3}{c}{COCOStuff validation split} \\
                \cmidrule{3-5}
                && $LPIPS\downarrow$ & $PSNR\uparrow$ & $SSIM\uparrow$ \\
                \midrule
                Box mask && 0.140 & 26.456 & 0.932 \\
                G.T. mask && 0.199 & 24.243 & 0.921 \\
                Fusion module && \textbf{0.110} & \textbf{28.592} & \textbf{0.944} \\
                \bottomrule
            \end{tabular}
        }
        \label{tab:ablation_mask}
    \end{subtable}
\end{table*}

\subsection{Ablation study}
\label{sec:ablation}
Here, we conduct ablation study to validate several important design choices in our model in \tabref{ablation}.
In all ablation study experiments, we use the COCO-Stuff validation dataset. 
First, we show that fusing features extracted from the instance network with the full-image network improve the performance.
Fusing features for both encoder and decoder perform the best. 
Second, we explore different strategies of selecting object bounding boxes as inputs for our instance network.
The results indicate that our default setting of choosing the top eight bounding boxes in terms of confidence score returned by object detector performs best and is slightly better than using the ground-truth bounding box.  
Third, we experiment with two alternative approaches (using the detected box as a mask or using the ground-truth instance mask provided in the COCO-Stuff dataset) for fusing features from multiple potentially overlapping object instances and the features from the full-image network.
Using our fusion module obtains a notable performance boost than the other two options.
This shows the capability of our fusion module to tackle more challenging scenarios with multiple overlapping objects.

\subsection{Runtime analysis}
Our colorization network involves two steps: 
(1) colorizing the individual instances and outputting the instance features; and 
(2) fusing the instance features into the full-image feature and producing a full-image colorization.
Using a machine with Intel i9-7900X 3.30GHz CPU, 32GB memory, and NVIDIA RTX 2080ti GPU, our average inference time over all the experiments is 0.187s for an image of resolution $256\times256$.
Each of two steps takes approximately $50\%$ of the running time, while the complexity of step 1 is proportional to the number of input instances and ranges from 0.013s (one instance) to 0.1s (eight instances).

\begin{figure}[!b]
    \mpage{0.31}{\includegraphics[width=\linewidth]{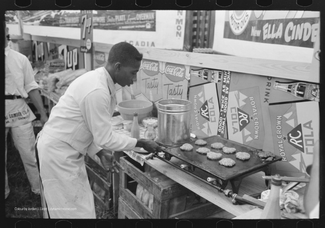}} \hfill
    \mpage{0.31}{\includegraphics[width=\linewidth]{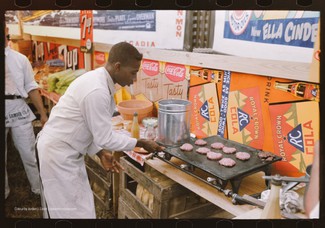}} \hfill
    \mpage{0.31}{\includegraphics[width=\linewidth]{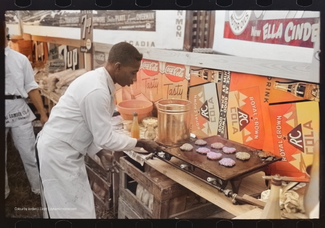}}
    \\
    \mpage{0.31}{\includegraphics[width=\linewidth]{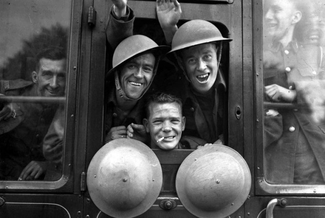}} \hfill
    \mpage{0.31}{\includegraphics[width=\linewidth]{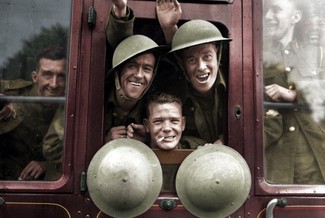}} \hfill
    \mpage{0.31}{\includegraphics[width=\linewidth]{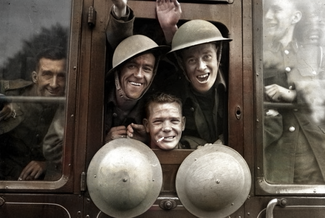}}
    \\
    \mpage{0.31}{\includegraphics[width=\linewidth]{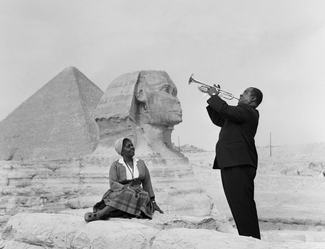}} \hfill
    \mpage{0.31}{\includegraphics[width=\linewidth]{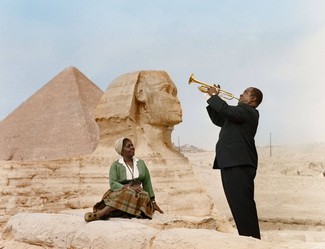}} \hfill
    \mpage{0.31}{\includegraphics[width=\linewidth]{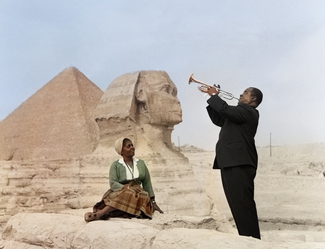}}
    \\
    \mpage{0.31}{Input} \hfill
    \mpage{0.31}{Expert} \hfill
    \mpage{0.31}{Our results}
    
    \caption{\tb{Colorizing legacy photographs.} 
    The middle column shows the manually colorized results by the experts.}
    \label{fig:legacy}
\end{figure}

\subsection{Colorizing legacy black and white photos}
\label{sec:legacy}
We apply our colorization model to colorize legacy black and white photographs. 
\figref{legacy} shows sample results along with manual colorization results by human expert\footnote{\url{bit.ly/color_history_photos}}.

\begin{figure}[!t]
    \begin{subfigure}[!t]{.34\linewidth}
       \includegraphics[width=\linewidth,keepaspectratio]{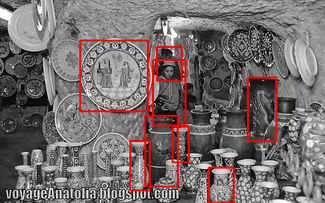}
       \includegraphics[width=\linewidth,keepaspectratio]{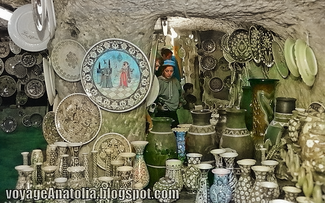}
       \caption{{Missing detections}}
    \end{subfigure}
    \begin{subfigure}[!t]{.65\linewidth}
       \includegraphics[width=.48\linewidth,keepaspectratio]{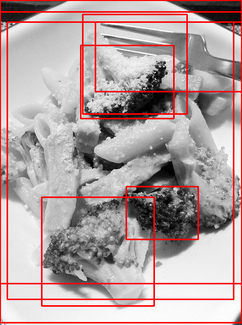}
       \includegraphics[width=.48\linewidth,keepaspectratio]{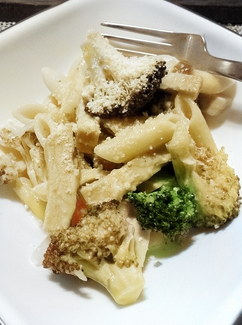}
       \caption{{Superimposed detections}}
    \end{subfigure}
    \caption{\tb{Failure cases.} (\emph{Left}) our model reverts back to the full-image colorization when a lot of vases are missing in the detection. (\emph{Right}) the fusion module may get confused when there are many superimposed object bounding boxes.
    }
    \label{fig:failure}
\end{figure}

\subsection{Failure modes}
\label{sec:failure}

We show 2 examples of failure cases in \figref{failure}. 
When the instances were not detected, our model reverts back to the full-image colorization network.
As a result, our method may produce visible artifacts such as washed-out colors or bleeding across object boundaries.

\ignorethis{
\begin{itemize}
    \item Imagenet~\cite{ILSVRC15}, as the reason describe in \secref{intro}, althogh a lot of existing work use Imagenet as train/test dataset, but it lacks of the instance diversity. 
    \begin{itemize}
        \item Training split: original train split, about 1.3 millon images
        \item Testing split: Follow the testing split proposed by~\cite{Larsson-ECCV-2016} ctest10k, Number of Images: 10000.
        \item All the method in the \tabref{comparision} are trained on Imagenet only.
        \item Quantitative comparisons (first column of \tabref{comparision}).
    \end{itemize}
    \item COCOStuff~\cite{caesar-CVPR-2018}, I think I should also put some figure here to show that COCOStuff do have more instance.
    \begin{itemize}
        \item Training images: 118k
        \item Label Data: bounding box, instance mask, and segmentation
        \item Thing (Instance) class: 80 classes
        \item More instance in single images.
        \item Testing Split: original validation set. 5000 images.
        \item Quantitative comparisons (second column of \tabref{comparision}).
    \end{itemize}
    \item Places205~\cite{zhou-CVPR-2014}, We also use this dataset to examinate every method's cross dataset generalization ability.
    \begin{itemize}
        \item Training images: 2.5 million images, 205 scene categories.
        \item Testing Images: 20500 images, 205 scene categories.
        \item Quantitative comparisons (third column of \tabref{comparision}).
    \end{itemize}
\end{itemize}

\heading{Evaluation metrics}
\begin{itemize}
    \item PSNR
    \item SSIM
    \item LPIPS Zhang~\etal~\cite{zhang-CVPR-2018}
    \begin{itemize}
        \item greate perceptual similarity
        \item Backbone: VGG as backbone
        \item Version: 0.1 %
    \end{itemize}
\end{itemize}

\heading{Training details}
Fill up the traingin detailes
\begin{itemize}
    \item machine specs:
    \begin{itemize}
        \item GPU: RTX 2080ti x 1
    \end{itemize}
    \item Based on the pre-trained weight provided by~\cite{Zhang-SIGGRAPH-2017}, so it don't need heavily training.
    \item Resolution: For all the input image, we resize to 256x256
    \item optimizer: Adam optimizer
    \begin{itemize}
        \item beta 1: 0.9
        \item beta 2: 0.999
    \end{itemize}
    \item 3 stage with different optimizer setting
    \begin{enumerate}
        \item train full image backbone
        \begin{itemize}
            \item epoch: 2
            \item learning rate: 0.00001
        \end{itemize}
        \item train instance image backbone
        \begin{itemize}
            \item epoch: 5
            \item learning rate:  0.00005
        \end{itemize}
        \item train fusion module
        \begin{itemize}
            \item epoch: 2
            \item learning rate: 0.00002
        \end{itemize}
    \end{enumerate}
    \item First train full image backbone, then use this weight and train instance image backbone then it could converge faster.
\end{itemize}

\subsection{Quantitative comparisons}\label{sec:quantitative}
\heading{Comparisons with baselines}
The quantitative comparison with the state-of-the-art is shown in \tabref{comparision}.
Provide two training weight:
\begin{itemize}
    \item Training on Imagenet: first six method in \tabref{comparision}
    \item Training on COCOStuff: the last two method with * sign in \tabref{comparision}
\end{itemize}

\heading{User study}
\begin{itemize}
    \item As shown in \figref{userstudy}, we evaluate our visual quality on 100 images with a forced-choice pairwise comparison.
    \item We will provide all the images used in user study in supplementary.
    \item Show some of our good case in user study and some bad case in user study.
\end{itemize}

\input{figure/fig_userstudy}

\subsection{Visual results}
\label{sec:visual}

\heading{Comparisons with the state-of-the-art}
We also show some of our cases and compare with other baselines (as shown in \figref{comparison}).

\begin{figure*}[!t]
    \begin{subfigure}[t]{.16\linewidth}
       \includegraphics[width=\linewidth,keepaspectratio]{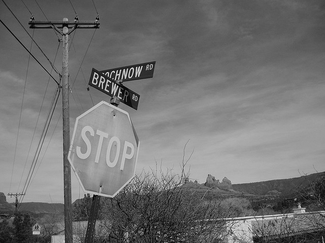}
       \includegraphics[width=\linewidth,keepaspectratio]{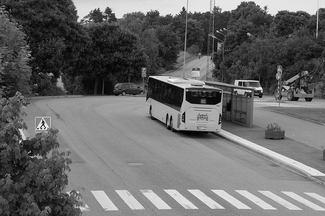}
       \includegraphics[width=\linewidth,keepaspectratio]{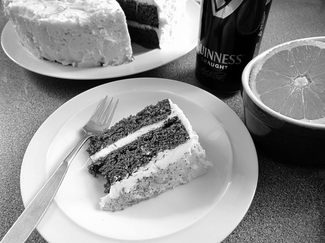}
       \includegraphics[width=\linewidth,keepaspectratio]{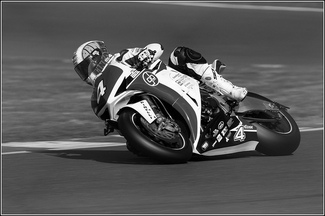}
       \includegraphics[width=\linewidth,keepaspectratio]{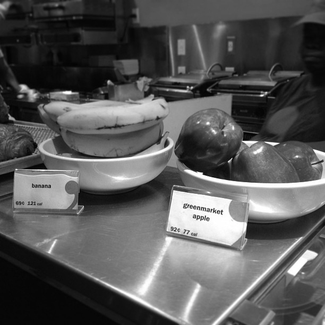}
       \caption{Input}
    \end{subfigure}
    \hfill
    \begin{subfigure}[t]{.16\linewidth}
       \includegraphics[width=\linewidth,keepaspectratio]{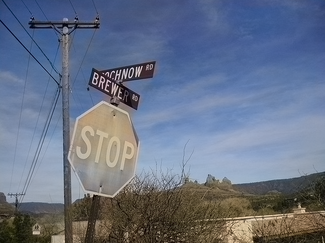}
       \includegraphics[width=\linewidth,keepaspectratio]{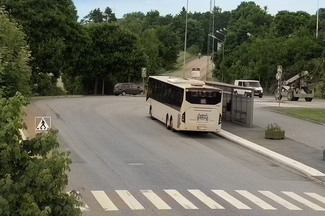}
       \includegraphics[width=\linewidth,keepaspectratio]{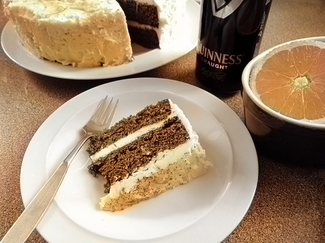}
       \includegraphics[width=\linewidth,keepaspectratio]{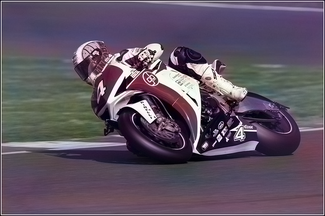}
       \includegraphics[width=\linewidth,keepaspectratio]{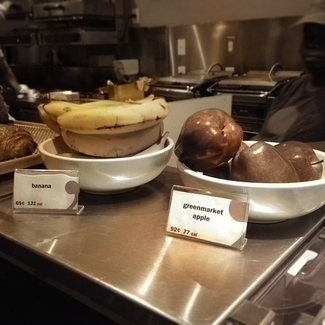}
       \caption{Iizuka~\etal~\cite{Iizuka-SIGGRAPH-2016}}
    \end{subfigure}
    \hfill
    \begin{subfigure}[t]{.16\linewidth}
       \includegraphics[width=\linewidth,keepaspectratio]{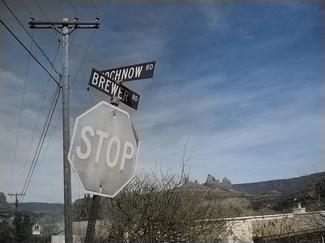}
       \includegraphics[width=\linewidth,keepaspectratio]{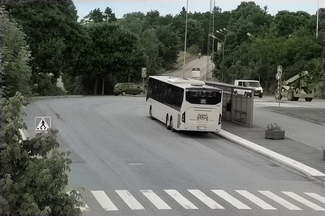}
       \includegraphics[width=\linewidth,keepaspectratio]{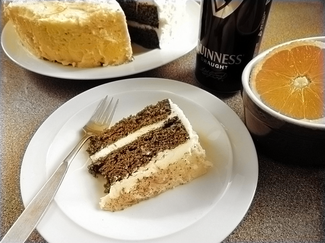}
       \includegraphics[width=\linewidth,keepaspectratio]{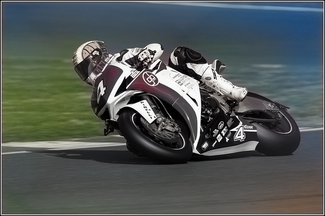}
       \includegraphics[width=\linewidth,keepaspectratio]{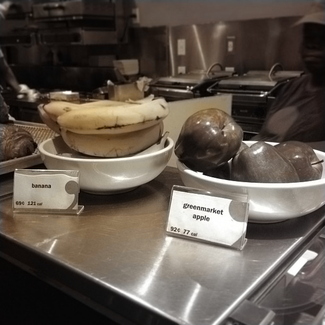}
       \caption{Larrson~\etal~\cite{larsson2017colorization}}
    \end{subfigure}
    \hfill
    \begin{subfigure}[t]{.16\linewidth}
       \includegraphics[width=\linewidth,keepaspectratio]{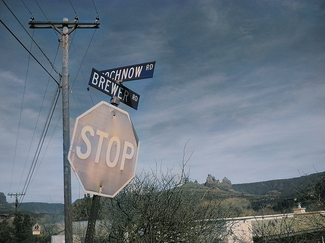}
       \includegraphics[width=\linewidth,keepaspectratio]{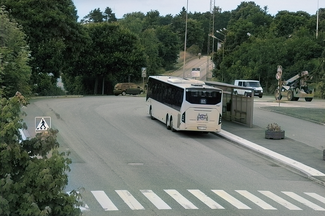}
       \includegraphics[width=\linewidth,keepaspectratio]{figure/images/visual_compare/000000496954_deoldify.png}
       \includegraphics[width=\linewidth,keepaspectratio]{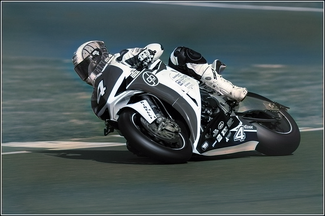}
       \includegraphics[width=\linewidth,keepaspectratio]{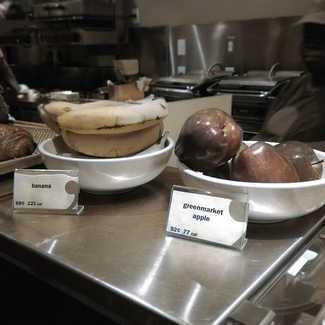}
       \caption{Deoldify~\cite{Deoldify}}
    \end{subfigure}
    \hfill
    \begin{subfigure}[t]{.16\linewidth}
       \includegraphics[width=\linewidth,keepaspectratio]{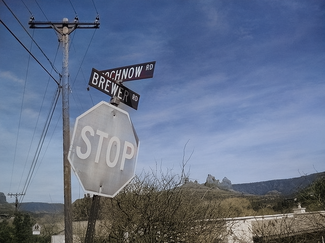}
       \includegraphics[width=\linewidth,keepaspectratio]{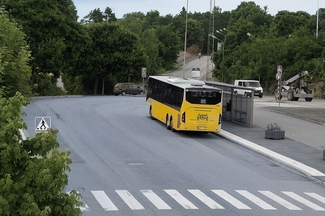}
       \includegraphics[width=\linewidth,keepaspectratio]{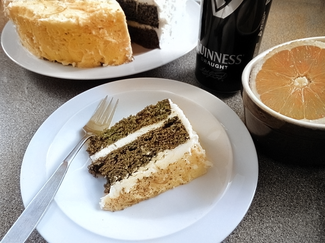}
       \includegraphics[width=\linewidth,keepaspectratio]{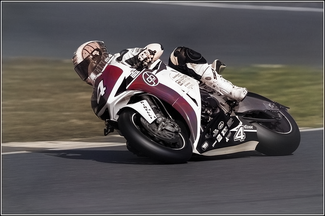}
       \includegraphics[width=\linewidth,keepaspectratio]{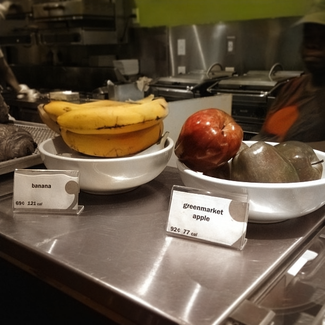}
       \caption{Zhang~\etal~\cite{Zhang-SIGGRAPH-2017}}
    \end{subfigure}
    \hfill
    \begin{subfigure}[t]{.16\linewidth}
       \includegraphics[width=\linewidth,keepaspectratio]{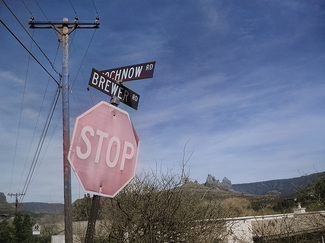}
       \includegraphics[width=\linewidth,keepaspectratio]{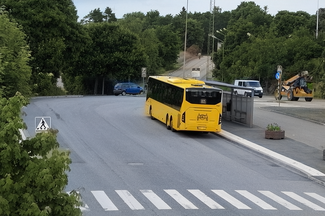}
       \includegraphics[width=\linewidth,keepaspectratio]{figure/images/visual_compare/000000496954_ours.png}
       \includegraphics[width=\linewidth,keepaspectratio]{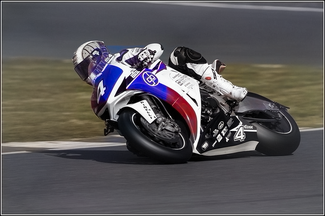}
       \includegraphics[width=\linewidth,keepaspectratio]{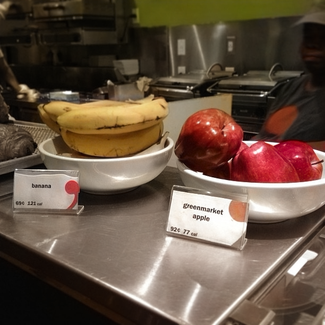}
       \caption{Ours}
    \end{subfigure}
    \caption{\tb{Visual Comparisons with the state-of-the-arts.} Our method predicts visually pleasing colors from complex scenes with multiple object instances.}
    \label{fig:comparison}
\end{figure*}

\heading{Visualizing the fusion network.}

We also visualize the the weighted mask in our different fusion module (as shown in \figref{vismask}).
\begin{figure}[!t]
    \begin{subfigure}[!t]{.39\linewidth}\centering
       \includegraphics[width=\linewidth,keepaspectratio]{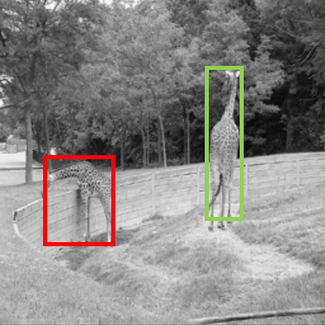}\\
       {Input}
    \end{subfigure}
    \begin{subfigure}[!t]{.19\linewidth}\centering
       \includegraphics[width=\linewidth,keepaspectratio]{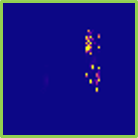}
       \includegraphics[width=\linewidth,keepaspectratio]{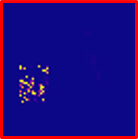}
       {Layer3}
    \end{subfigure}
    \begin{subfigure}[!t]{.19\linewidth}\centering
       \includegraphics[width=\linewidth,keepaspectratio]{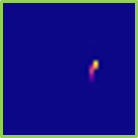}
       \includegraphics[width=\linewidth,keepaspectratio]{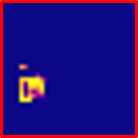}\\
       {Layer7}
    \end{subfigure}
    \begin{subfigure}[!t]{.19\linewidth}\centering
       \includegraphics[width=\linewidth,keepaspectratio]{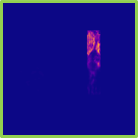}
       \includegraphics[width=\linewidth,keepaspectratio]{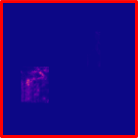}\\
       {Layer10}
    \end{subfigure}
    \\
    \begin{subfigure}[!t]{0.49\linewidth}\centering
        \includegraphics[width=\linewidth,keepaspectratio]{figure/images/vismask/mask_fig_UG.png}
        {Zhang~\etal~\cite{Zhang-SIGGRAPH-2017}}
    \end{subfigure}
    \begin{subfigure}[!t]{0.49\linewidth}\centering
        \includegraphics[width=\linewidth,keepaspectratio]{figure/images/vismask/mask_fig_ours.png}
        {Our results}
    \end{subfigure}
    \caption{\tb{Visualizing the fusion network.} The visualized weighted mask in layer3, layer7 and layer10 show that our model learns to \emph{adaptively} blend the features across different layers. Fusing instance-level features help improve colorization.}
    \label{fig:vismask}
\end{figure}

\subsection{Ablation study}
\label{sec:ablation}

\begin{itemize}
    \item To show that fusing all the layers instance network with full image network would be the best, we evaluate different fusion parts in our network and shown in \tabref{ablation}.
    \item To show that as the number of the bounding boxes or the accuracy of the boxes' position increase, we could get better results and shown in \tabref{ablation_box}.
    \item To show that our weight mask is the good fusing strategy, we do some evaluate on different fusing method and shown in \tabref{ablation_mask}.
\end{itemize}

\begin{figure}[!b]
    \mpage{0.31}{\includegraphics[width=\linewidth]{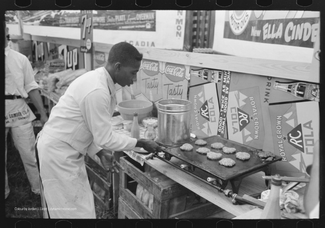}} \hfill
    \mpage{0.31}{\includegraphics[width=\linewidth]{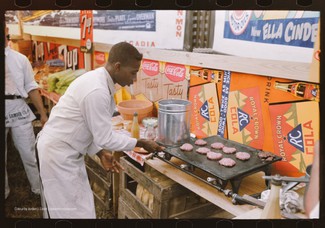}} \hfill
    \mpage{0.31}{\includegraphics[width=\linewidth]{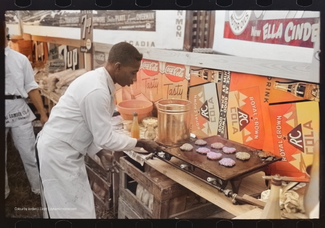}}
    \\
    \mpage{0.31}{\includegraphics[width=\linewidth]{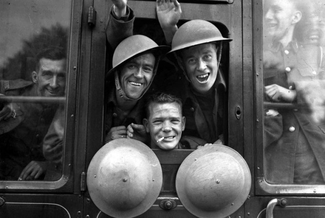}} \hfill
    \mpage{0.31}{\includegraphics[width=\linewidth]{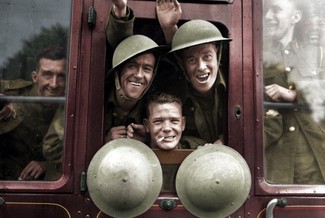}} \hfill
    \mpage{0.31}{\includegraphics[width=\linewidth]{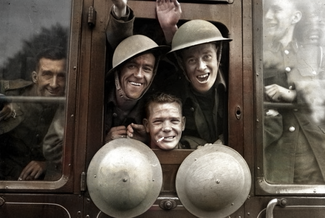}}
    \\
    \mpage{0.31}{\includegraphics[width=\linewidth]{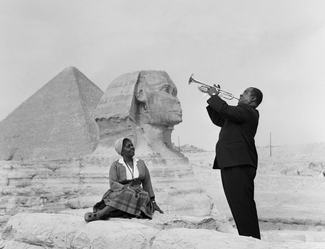}} \hfill
    \mpage{0.31}{\includegraphics[width=\linewidth]{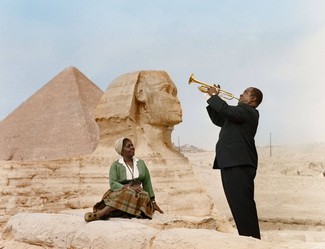}} \hfill
    \mpage{0.31}{\includegraphics[width=\linewidth]{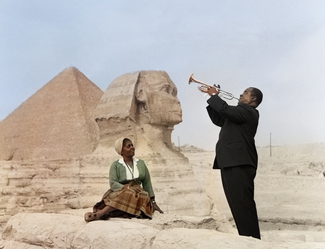}}
    \\
    \mpage{0.31}{Input} \hfill
    \mpage{0.31}{Expert} \hfill
    \mpage{0.31}{Our results} \\
    \caption{\tb{Colorizing legacy photographs.} 
    The middle column shows the manually colorized results by the experts.}
    \label{fig:legacy}
\end{figure}

\subsection{Colorizing legacy black and white photos}
\label{sec:legacy}
For the real world legacy images, we can still have some good results on instance part (as shown in \figref{legacy}).

\subsection{Diverse colorization}
\label{sec:diverseColorization}
\begin{itemize}
    \item Showing that our architecture could be easily applied to other existing models and produce the reasonable results.
    \item Show some results with static background color, but diverse color on instance part.
\end{itemize}

\subsection{Failure modes}
\label{sec:failure}
\begin{itemize}
    \item failure from object detection failed on some object
    \item failed because of some other reason
\end{itemize}
}

\section{Conclusions}
\label{sec:conclusions}
We present a novel instance-aware image colorization. By leveraging an off-the-shelf object detection model to crop out the images, our architecture extracts the feature from our instance branch and full images branch, then we fuse them with our newly proposed fusion module and obtain a better feature map to predict the better results. Through extensive experiments, we show that our work compares favorably against existing methods on three benchmark datasets.

\paragraph{Acknowledgements.} The project was funded in part by the Ministry of Science and Technology of Taiwan (108-2218-E-007 -050- and 107-2221-E-007-088-MY3).

{\small
\bibliographystyle{ieee_fullname}
\bibliography{main}
}

\onecolumn
\appendix

\title{Instance-aware Image Colorization\\Supplemental Material}
\author{}
\maketitle

In this supplementary document, we provide additional visual comparisons and quantitative evaluation to complement the main manuscript.

\section{Visualization of Fusion Module}
We show two images where multiple instances have been detected by the object detection model.
We visualize the weighted mask predicted by our fusion module at multiple layers (3rd, 7th, and 10th layers).
Note that our fusion module learns to \emph{adaptively} blend the features extracted from the instance colorization branch to enforce coherent colorization for the entire image.

\begin{figure}[h]
    \begin{subfigure}[!t]{.19\linewidth}\centering
       \includegraphics[width=\linewidth,keepaspectratio]{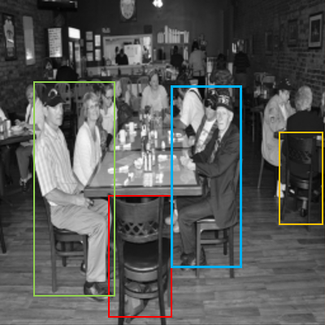}\\
       {Input}
    \end{subfigure}
    \hfill
    \begin{subfigure}[!t]{.09\linewidth}\centering
       \includegraphics[width=\linewidth,keepaspectratio]{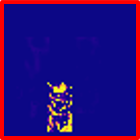}
       \includegraphics[width=\linewidth,keepaspectratio]{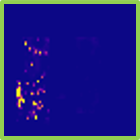}\\
       {Layer3}
    \end{subfigure}
    \hfill
    \begin{subfigure}[!t]{.09\linewidth}\centering
       \includegraphics[width=\linewidth,keepaspectratio]{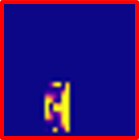}
       \includegraphics[width=\linewidth,keepaspectratio]{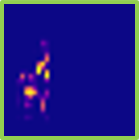}\\
       {Layer7}
    \end{subfigure}
    \hfill
    \begin{subfigure}[!t]{.09\linewidth}\centering
       \includegraphics[width=\linewidth,keepaspectratio]{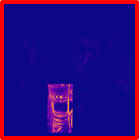}
       \includegraphics[width=\linewidth,keepaspectratio]{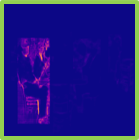}\\
       {Layer10}
    \end{subfigure}
    \hfill
    \begin{subfigure}[!t]{.09\linewidth}\centering
       \includegraphics[width=\linewidth,keepaspectratio]{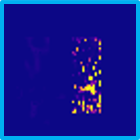}
       \includegraphics[width=\linewidth,keepaspectratio]{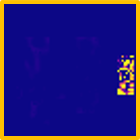}\\
       {Layer3}
    \end{subfigure}
    \hfill
    \begin{subfigure}[!t]{.09\linewidth}\centering
       \includegraphics[width=\linewidth,keepaspectratio]{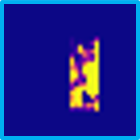}
       \includegraphics[width=\linewidth,keepaspectratio]{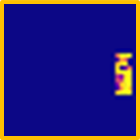}\\
       {Layer7}
    \end{subfigure}
    \hfill
    \begin{subfigure}[!t]{.09\linewidth}\centering
       \includegraphics[width=\linewidth,keepaspectratio]{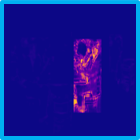}
       \includegraphics[width=\linewidth,keepaspectratio]{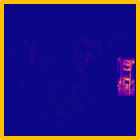}\\
       {Layer10}
    \end{subfigure}
    \hfill
    \begin{subfigure}[!t]{.19\linewidth}\centering
       \includegraphics[width=\linewidth,keepaspectratio]{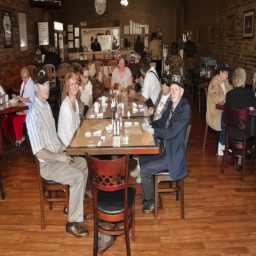}\\
       {Output}
    \end{subfigure}
    \begin{subfigure}[!t]{.19\linewidth}\centering
       \includegraphics[width=\linewidth,keepaspectratio]{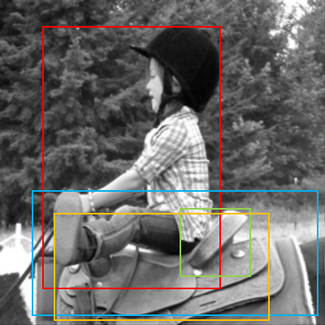}\\
       {Input}
    \end{subfigure}
    \hfill
    \begin{subfigure}[!t]{.09\linewidth}\centering
       \includegraphics[width=\linewidth,keepaspectratio]{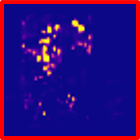}
       \includegraphics[width=\linewidth,keepaspectratio]{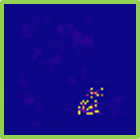}\\
       {Layer3}
    \end{subfigure}
    \hfill
    \begin{subfigure}[!t]{.09\linewidth}\centering
       \includegraphics[width=\linewidth,keepaspectratio]{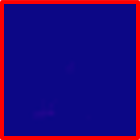}
       \includegraphics[width=\linewidth,keepaspectratio]{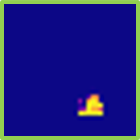}\\
       {Layer7}
    \end{subfigure}
    \hfill
    \begin{subfigure}[!t]{.09\linewidth}\centering
       \includegraphics[width=\linewidth,keepaspectratio]{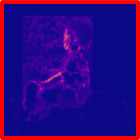}
       \includegraphics[width=\linewidth,keepaspectratio]{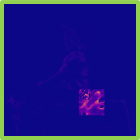}\\
       {Layer10}
    \end{subfigure}
    \hfill
    \begin{subfigure}[!t]{.09\linewidth}\centering
       \includegraphics[width=\linewidth,keepaspectratio]{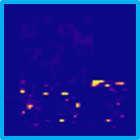}
       \includegraphics[width=\linewidth,keepaspectratio]{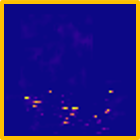}\\
       {Layer3}
    \end{subfigure}
    \hfill
    \begin{subfigure}[!t]{.09\linewidth}\centering
       \includegraphics[width=\linewidth,keepaspectratio]{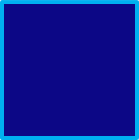}
       \includegraphics[width=\linewidth,keepaspectratio]{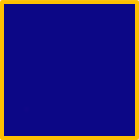}\\
       {Layer7}
    \end{subfigure}
    \hfill
    \begin{subfigure}[!t]{.09\linewidth}\centering
       \includegraphics[width=\linewidth,keepaspectratio]{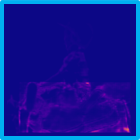}
       \includegraphics[width=\linewidth,keepaspectratio]{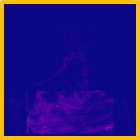}\\
       {Layer10}
    \end{subfigure}
    \hfill
    \begin{subfigure}[!t]{.19\linewidth}\centering
       \includegraphics[width=\linewidth,keepaspectratio]{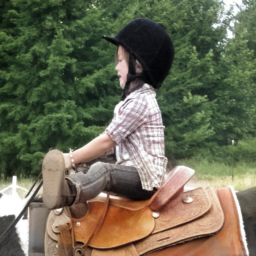}\\
       {Output}
    \end{subfigure}
    \caption{\tb{Visualizing the fusion network.} The visualized weighted mask in layer3, layer7 and layer10.}
    \label{fig:vismask_supp}
\end{figure}

\section{Extended quantitative evaluation}
In this section, we provide two additional quantitative evaluation. 

\heading{Baselines using pre-trained weights.}
As we mentioned in our paper, we use Zhang et al.~\cite{Zhang-SIGGRAPH-2017} as our backbone colorization model. However, the model in~\cite{Zhang-SIGGRAPH-2017} is trained for guidance colorization with a resolution of 176$\times$176. In our setting, we need a fully-automatic colorization model with resolution 256$\times$256. In light of this, we retrain the model on the ImageNet dataset to fit our setting. Here we carry out an experiment on the COCOStuff dataset.
We show a quantitative comparison and qualitative comparison between (1) the original pre-trained weight provided by the authors and (2) the weight after the fine-tuning step.
In general, re-training the model under the resolution of 256$\times$256 results in slight performance degradation under PSNR and similar performance in terms of LPIPS and SSIM with respect to the original pre-trained model.

\begin{table}[h]
    \centering
    \caption{\textbf{Quantitative Comparison with different models weight on COCOStuff} We test our colorization backbone model~\cite{Zhang-SIGGRAPH-2017} with different weight on COCOStuff.}
    \resizebox{.5\textwidth}{!}{
        \begin{tabular}{@{}lclll@{}}
            \toprule
            \multirow{2}{*}{Method} & \phantom{abc} & 
            \multicolumn{3}{c}{COCOStuff validation split} \\
            \cmidrule{3-5}
            && $LPIPS\downarrow$ & $PSNR\uparrow$ & $SSIM\uparrow$ \\
            \midrule
            (a) Original model weight~\cite{Zhang-SIGGRAPH-2017} && 0.133 & 27.050 & 0.937 \\
            (b) Retrain on the ImageNet dataset && 0.138 & 26.823 & 0.937 \\
            (c) Finetuned on the COCOStuff dataset && \textbf{0.128} & \textbf{27.251} & \textbf{0.938} \\
            \bottomrule
        \end{tabular}
    }
    \label{tab:weight_comp}
\end{table}

\begin{figure}[h]
    \centering
    \begin{subfigure}[t]{.24\linewidth}\centering
       \includegraphics[width=\linewidth,keepaspectratio]{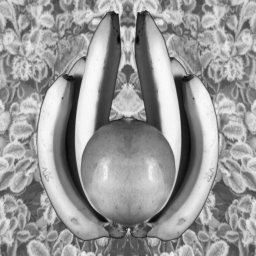}\\
       \includegraphics[width=\linewidth,keepaspectratio]{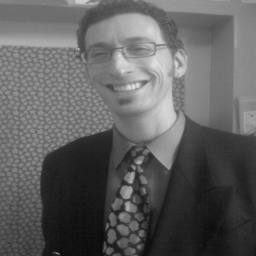}\\
       \caption{(Gray)}
    \end{subfigure}
    \begin{subfigure}[t]{.24\linewidth}\centering
       \includegraphics[width=\linewidth,keepaspectratio]{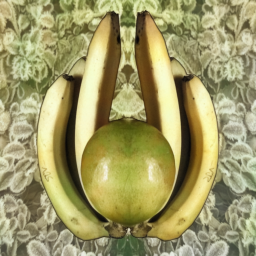}\\
       \includegraphics[width=\linewidth,keepaspectratio]{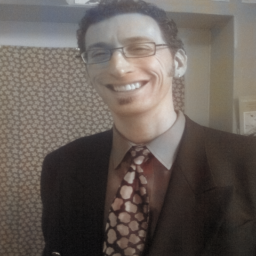}\\
       \caption{\centering Original model weight~\cite{Zhang-SIGGRAPH-2017}}
    \end{subfigure}
    \begin{subfigure}[t]{.24\linewidth}\centering
       \includegraphics[width=\linewidth,keepaspectratio]{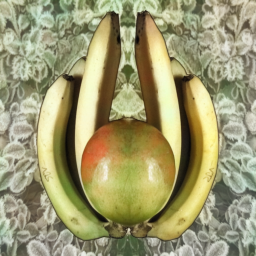}\\
       \includegraphics[width=\linewidth,keepaspectratio]{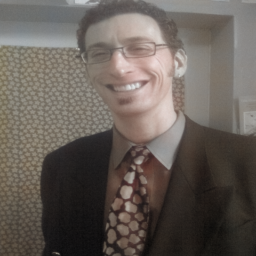}\\
       \caption{\centering Retrain on the ImageNet dataset}
    \end{subfigure}
    \begin{subfigure}[t]{.24\linewidth}\centering
       \includegraphics[width=\linewidth,keepaspectratio]{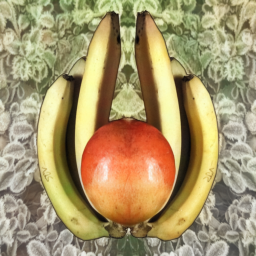}\\
       \includegraphics[width=\linewidth,keepaspectratio]{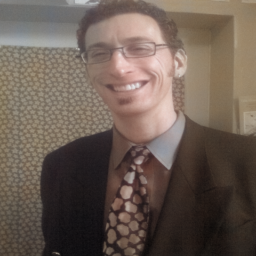}\\
       \caption{\centering Finetuned on the COCOStuff dataset}
    \end{subfigure}
    \captionof{figure}{\tb{Different model weight of~\cite{Zhang-SIGGRAPH-2017}} We present some images that are inferencing from different weight of~\cite{Zhang-SIGGRAPH-2017}. The images above from left to right is Grayscale, Original weight, retrain on Imagenet, and fine-tuned on COCOStuff.
    }
    \label{fig:diffw}
\end{figure}

\section{User Study setup}
Here we describe the detailed procedure of our user study.
We first ask the subjects to read a document, including the instruction of the webpage and selection basis (according to the color correctness and naturalness).
The subjects then start a pair-wise forced-choice \emph{without} any ground truth reference or grayscale reference (i.e., no-reference tests). 
We embed two redundant comparisons for sanity check (i.e the same comparisons appear in the user study twice).
We reject the votes if the subjects do not answer consistently for both redundant comparisons.
We summarize the results in the main paper. 
We provide all the images used in our user study as well as the user preference votes in the supplementary web-based viewer (click the `User Study Result' button). 

\section{Failure cases}
\begin{figure}[h]
    \centering
    \begin{subfigure}[!t]{.28\linewidth}\centering
       \includegraphics[width=\linewidth,keepaspectratio]{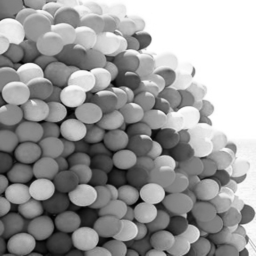}\\
       \includegraphics[width=\linewidth,keepaspectratio]{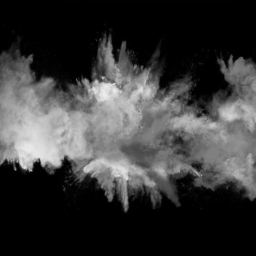}\\
       \caption{Grayscale input}
    \end{subfigure}
    \begin{subfigure}[!t]{.28\linewidth}\centering
       \includegraphics[width=\linewidth,keepaspectratio]{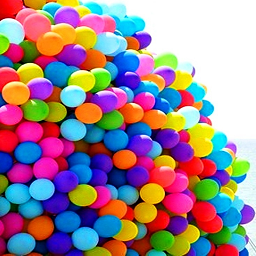}\\
       \includegraphics[width=\linewidth,keepaspectratio]{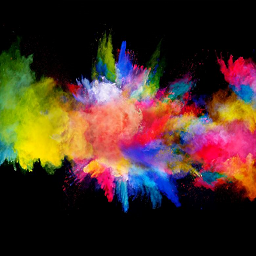}\\
       \caption{Ground truth color image}
    \end{subfigure}
    \begin{subfigure}[!t]{.28\linewidth}\centering
       \includegraphics[width=\linewidth,keepaspectratio]{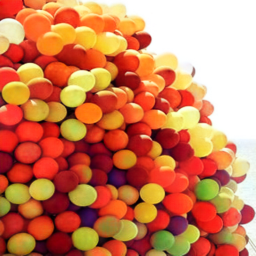}\\
       \includegraphics[width=\linewidth,keepaspectratio]{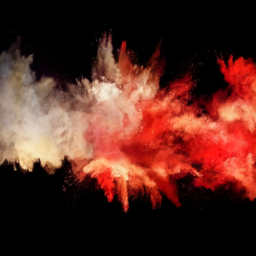}\\
       \caption{Our result}
    \end{subfigure}
    \captionof{figure}{\tb{Failure cases.} We present some images that are out of distribution and are capable of colorizing plausible color. The images above from left to right is (a) grayscale input, (b) ground truth color image and (c) our result.
    }
    \label{fig:degenerated}
\end{figure}

While we have shown significantly improved results, single image colorization remains a challenging problem. 
\figref{degenerated} shows two examples where our model is unable to predict the vibrant, bright colors from the given grayscale input image. 

\end{document}